\documentclass[runningheads]{llncs}
\usepackage{amsmath,amssymb} 
\usepackage{color}
\usepackage{graphicx}
\usepackage{url}
\newcommand{\etal}{\textit{et al}.}

\begin{document}

\title{Long-Term Visual Object Tracking Benchmark} 
\titlerunning{Long-Term Visual Object Tracking Benchmark} 


\author{Abhinav Moudgil\and
Vineet Gandhi}
%

\authorrunning{A. Moudgil and V. Gandhi} 


\institute{Center for Visual Information Technology, Kohli Center on Intelligent Systems
International Institute of Information Technology, Hyderabad, India
\email{abhinav.moudgil@research.iiit.ac.in, vgandhi@iiit.ac.in}}

\maketitle

\begin{abstract}
We propose a new long video dataset\footnote{Dataset and tracking results are available at \url{https://amoudgl.github.io/tlp/}} (called Track Long and Prosper  - TLP) and benchmark for single object tracking. The dataset consists of 50 HD videos from real world scenarios, encompassing a duration of over 400 minutes (676K frames), making it more than 20 folds larger in average duration per sequence and more than 8 folds larger in terms of total covered duration, as compared to existing generic datasets for visual tracking. The proposed dataset paves a way to suitably assess long term tracking performance and train better deep learning architectures (avoiding/reducing augmentation, which may not reflect real world behaviour). We benchmark the dataset on 17 state of the art trackers and rank them according to tracking accuracy and run time speeds. We further present thorough qualitative and quantitative evaluation highlighting the importance of long term aspect of tracking. Our most interesting observations are (a) existing short sequence benchmarks fail to bring out the inherent differences in tracking algorithms which widen up while tracking on long sequences and (b) the accuracy of trackers abruptly drops on challenging long sequences, suggesting the potential need of research efforts in the direction of long-term tracking.
\end{abstract}
\section{Introduction}
Visual tracking is a fundamental task in computer vision and is a key component in wide range of applications like surveillance, autonomous navigation, video analysis and editing, augmented reality etc. Many of these applications rely on long-term tracking, however, only few tracking algorithms have focused on the challenges specific to long duration aspect~\cite{kalal2012tracking,ma2015long,hua2014,supancic2013self}. Although they conceptually attack the long term aspect, the evaluation is limited to shorter sequences or couple of selected longer videos. The recent correlation filter~\cite{DanelljanCVPR2017,DanelljanECCV2016,Bertinetto_2016_CVPR,zhang2017multi} and deep learning~\cite{yun2017adnet,nam2016mdnet,bertinetto2016fully,held2016learning} based approaches have significantly advanced the field, however, their long term applicability is also unapparent as the evaluation is limited to datasets with typical average video duration of about 20-40 seconds. Not just the evaluation aspect, the lack of long term tracking datasets has been a hindrance for training in several recent state of the art approaches. These methods either limit themselves to available small sequence data~\cite{nam2016mdnet,yun2017adnet} or use augmentation on datasets designed for other tasks like object detection~\cite{held2016learning}. 

Motivated by the above observation, we propose a new long duration dataset called Track Long and Prosper (TLP), consisting of 50 long sequences. The dataset covers a wide variety of target subjects and is arguably one of the most challenging datasets in terms of occlusions, fast motion, viewpoint change, scale variations etc. However, compared to existing generic datasets, the most prominent aspect of TLP dataset is that it is larger by more than 20 folds in terms of average duration per sequence, which makes it ideal to study challenges specific to long duration aspect. For example, drift is a common problem in several tracking algorithms and it is not always abrupt and may occur due to accumulation of error over time (which may be a slow procedure and can be difficult to gauge in short sequences). Similarly, long sequences allow us to study the consistency of a tracker to recover from momentary failures. 

We select 17 recent state of the art trackers which are scalable to be evaluated on TLP dataset and provide a thorough evaluation in terms of tracking accuracy and real time performance. Testing on such a large dataset significantly reduces the overfitting problem, if any, and reflects if the tracker is actually designed to consistently recover from challenging scenarios. To present a further perspective, we provide a comprehensive attribute wise comparison of different tracking algorithms by selecting various sets of short sequences (derived from original TLP sequences), in which each set only contains sequences where a particular type of challenge is dominant (like illumination variation, occlusions, out of view etc.).

We observe that the rankings from previous short sequence datasets like OTB50~\cite{wu2013online} significantly vary from the rankings obtained on the proposed TLP dataset. Several top ranked trackers on recent benchmarks fail to adapt to long-term tracking scenario and their performance drops significantly. Additionally, the performance margin notably widens among several trackers, whose performances are imperceptibly close in existing benchmarks. More specifically, apart from MDNet~\cite{nam2016mdnet}, performance of all other evaluated tracker drops below 25\% on commonly used metric of area under the curve of success plots. Our investigation hence strongly highlights the need for more research efforts in long term tracking and to our knowledge the proposed dataset and benchmark is the first systematic exploration in this direction.

\section{Related Work}
\subsection{Tracking Datasets}
\begin{table}
\label{table:comparison}
\caption{Comparing TLP with other object tracking datasets.}
\scriptsize
\scalebox{0.9}{
\begin{tabular}{l @{\hskip 2ex} c @{\hskip 2ex} c @{\hskip 2ex} c @{\hskip 2ex} c @{\hskip 2ex} c @{\hskip 2ex} c}
\hline
& Frame rate  & \# videos & Min Duration & Mean Duration & Max Duration & Total Duration\\
& (FPS)  &   & (sec) & (sec) & (sec) & (sec)\\ \hline
  UAV123\cite{mueller2016benchmark}  & 30  & 123 &  3.6  & 30.5  & 102.8  & 3752  \\
  OTB50\cite{wu2013online}  & 30  & 51 &  2.3  & 19.3  & 129  & 983  \\
  OTB100\cite{wu2015object}  & 30  & 100  & 2.3    & 19.6   & 129  & 1968  \\
  TC128\cite{liang2015encoding}   & 30  & 129  & 2.3   & 14.3  & 129  & 1844 \\
  VOT14\cite{kristan2014visual}   & 30  & 25 & 5.7   &  13.8 & 40.5  &  346 \\
  VOT15\cite{kristan2015visual}   & 30  & 60 & 1.6   & 12.2  & 50.2  & 729 \\
  ALOV300\cite{smeulders2014visual}   & 30  & 314 & 0.6  & 9.2  & 35  & 2978 \\
    NFS\cite{galoogahi2017need}   & 240  & 100 &0.7 & 16  & 86.1  & 1595 \\
{\bf TLP}   & 24/30  & 50 & {\bf 144} & {\bf 484.8}  & {\bf 953}  & {\bf 24240} \\
\hline 
\end{tabular}
}

\end{table}

There are several existing datasets which are widely used for evaluating the tracking algorithms and are summarized in Table~\ref{table:comparison}. The OTB50~\cite{wu2013online}, OTB100~\cite{wu2015object} are the most commonly used ones. They include 50 and 100 sequences respectively and capture a generic real world scenario (where some videos are taken from platforms like YouTube and some are specifically recorded for tracking application). They provide per frame bounding box annotation and per sequence annotation of attributes like illumination variation, occlusion, deformation etc. 

The ALOV300++ dataset~\cite{smeulders2014visual} focuses on diversity and includes more than 300 short sequences (average length of only about 9 seconds). The annotations in ALOV300++ dataset are made every fifth frame. A small set of challenging sequences (partially derived from OTB50, OTB100 and ALOV300++ datasets) has been used in VOT14~\cite{kristan2014visual} and VOT15~\cite{kristan2015visual} datasets. They extend the rectangular annotations to rotated ones and provide per frame attribute annotations, for more accurate evaluation. Both of these datasets have been instrumental in yearly visual object tracking (VOT) challenge. 

Some datasets have focused on particular type of applications/aspects. TC128 \cite{liang2015encoding} was proposed to study the role of color information in tracking. It consists of 128 sequences (some of them are common to OTB100 dataset) and provides per frame annotations and sequence wise attributes. Similarly, UAV~\cite{mueller2016benchmark} targets the tracking application, when the videos are captured from low-altitude unmanned aerial vehicles. 
The focus of their work is to highlight challenges incurred while tracking in video taken from an aerial viewpoint. They provide both real and synthetically generated UAV videos with per frame annotations.

More recently, two datasets were proposed to incorporate the benefits of advances in capture technology. The NFS~\cite{galoogahi2017need} dataset was proposed to study the fine grained variations in tracking by capturing high frame rate videos (240 FPS). Their analysis shows that since high frame video reduces appearance variation per frame, it is possible to achieve state of the art performance using substantially simpler tracking algorithms. Another recent dataset called AMP~\cite{cehovinbeyond}, explores the utility of 360$^{\circ}$ videos to generate and study tracking with typical motion patterns (which can be achieved by varying the camera re-parametrization in omni-directional videos). Contemporary to our work,~\cite{lukevzivc2018now} and \cite{valmadre2018long} also review recent trackers for long-term tracking. However, they limit the long-term tracking definition to the ability of a tracker to re-detect after object goes out of view and the quality of their long term datasets is lower than our proposed TLP dataset, in terms of resolution and per sequence length. We evaluate the trackers from a holistic perspective and show that even if there is no apparent major challenge or target disappearance, tracking consistently for a long period of time is an extremely challenging task.

Although recent advances pave the way to explore several novel and specific fine grained aspects, the crucial long term tracking aspect is still missing from most of the current datasets. The typical average length per sequence is still only about 10-30 seconds. The proposed TLP dataset takes it to about 8-9 minutes per sequence, making it the largest densely annotated high-resolution dataset for the application of visual object tracking.
\subsection{Tracking Methods}
Most of the earlier approaches trained a model/classifier considering the initial bounding box marked by the user as ``foreground" and areas farther away from the annotated box as ``background". The major challenge in most of these approaches is to properly update the model or the classifier over time to reduce drift. This problem has been tackled in several innovative ways, such as Multiple Instance Learning~\cite{babenko2011robust}, Online Boosting~\cite{grabner2008semi}, P-N Learning~\cite{kalal2012tracking}, or by using ensemble of classifiers trained/initiated at difference instances of time~\cite{zhang2014meem}. Most of the recent advances, however, have focused only in two major directions i.e Correlation Filter (CF) based tracking~\cite{DanelljanCVPR2017,DanelljanECCV2016,Bertinetto_2016_CVPR,zhang2017multi} and deep learning based tracking~\cite{yun2017adnet,nam2016mdnet,bertinetto2016fully,held2016learning}. 
The CF based trackers have gained huge attention due to their computational efficiency derived by operating in Fourier domain and the capability of efficient online adaptation. The interest in CF based approaches was kindled by the MOSSE~\cite{bolme2010visual} tracker proposed by Bolme \etal, which demonstrated an impressive speed of about 700 FPS. Thereafter, several works have built upon this idea and have significantly improved tracking accuracy. The list includes ideas of using kernelized correlation filters~\cite{henriques2012exploiting}; exploiting multi-dimensional features~\cite{danelljan2014adaptive,henriques2015high}; combining template based features with pixel wise information for robustness to deformation and scale variations~\cite{li2014scale,Bertinetto_2016_CVPR}; employing kernel ridge regression to reduce drift~\cite{ma2015long} etc. The work by Kiani \etal~\cite{kiani2015correlation} identified the boundary effects in Fourier domain as one of the reasons for sub-optimal performance of CF based approaches. Solutions such as Spatially regularized CF~\cite{danelljan2015learning} and Background aware CF~\cite{kiani2017learning} were later proposed to mitigate the boundary effects.  

More recent efforts in CF based trackers utilize deep convolutional features instead of hand crafted ones like HOG~\cite{dalal2005histograms}. Multiple convolutional layers in a hierarchical ensemble of independent DCF trackers was employed by Ma \etal~\cite{Ma-ICCV-2015}. Danelljan \etal~\cite{DanelljanECCV2016} extended it by fusing multiple convolutional layers with different spatial resolutions in a joint learning framework. This combination of deep CNN features has led CF based trackers to the top of several benchmarks like OTB50; however, they come with an additional computational cost. Some recent efforts have been made to enhance running speeds by using ideas like factorized convolutions~\cite{DanelljanCVPR2017}; however, the speeds are still much slower than the traditional CF trackers. 

\begin{figure}
        \centering
        \setlength{\tabcolsep}{0pt}
        \renewcommand{\arraystretch}{0}
        \begin{tabular}{ccccc} 
        \includegraphics[width=0.2\linewidth]{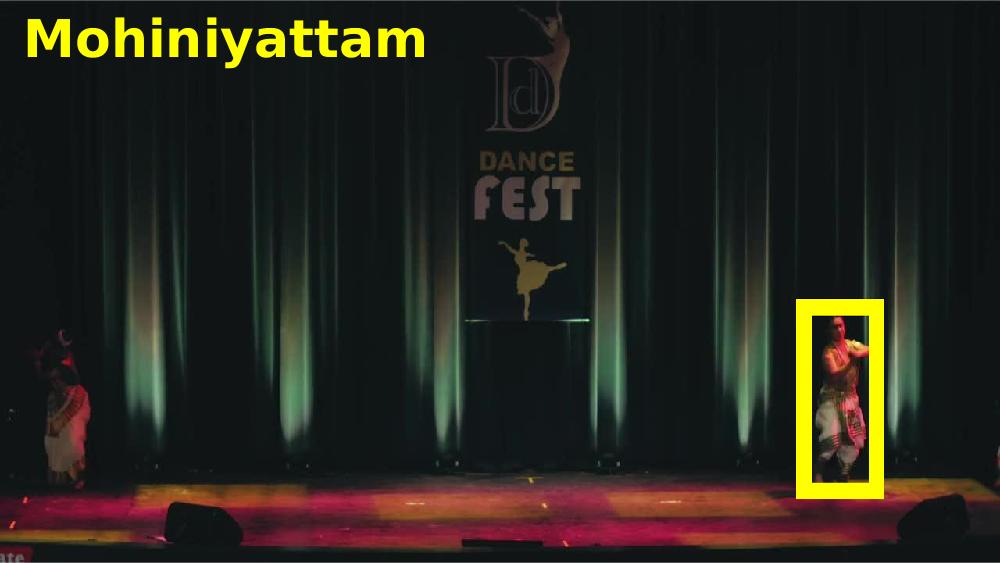}& 
        \includegraphics[width=0.2\linewidth]{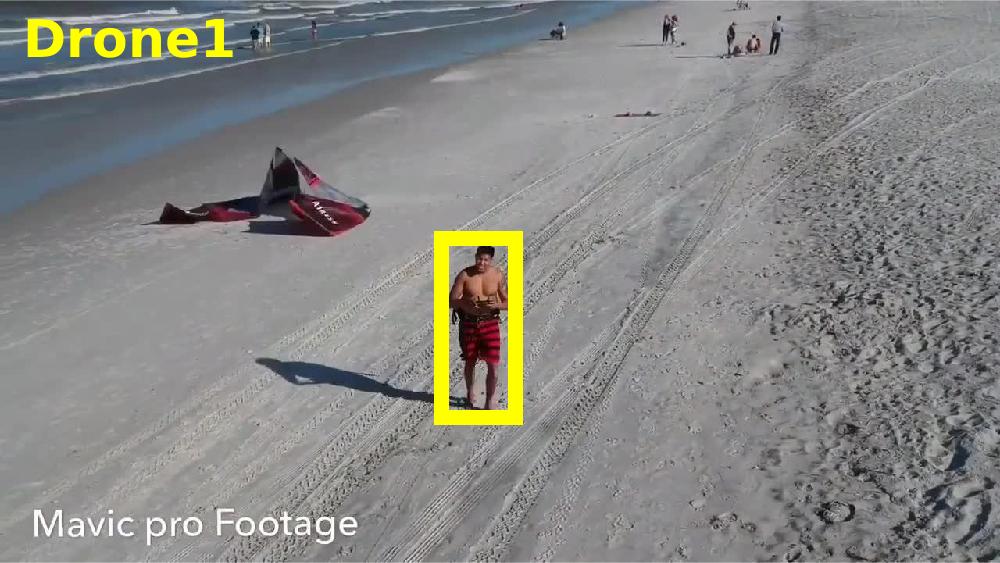}& 
        \includegraphics[width=0.2\linewidth]{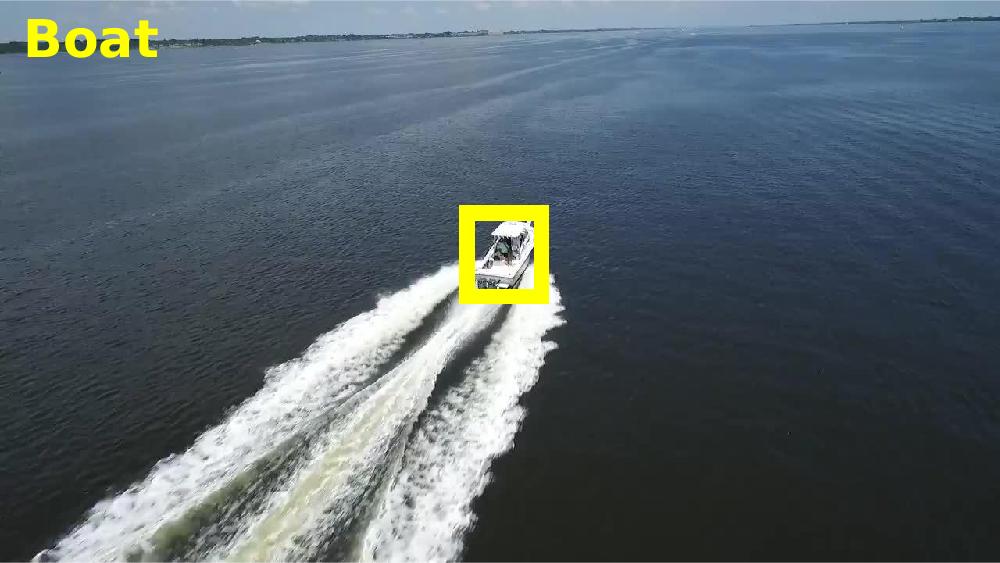}&
        \includegraphics[width=0.2\linewidth]{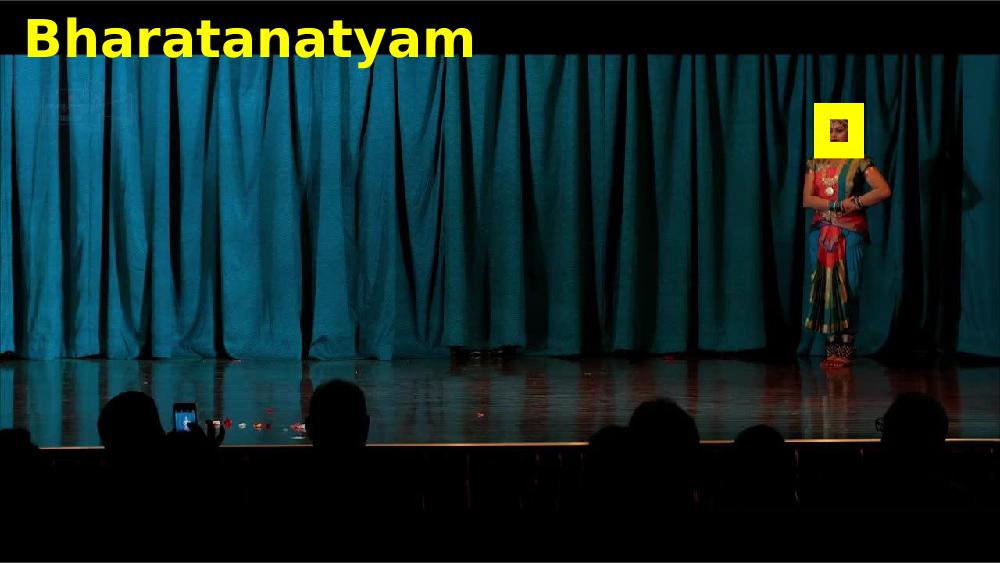}& 
        \includegraphics[width=0.2\linewidth]{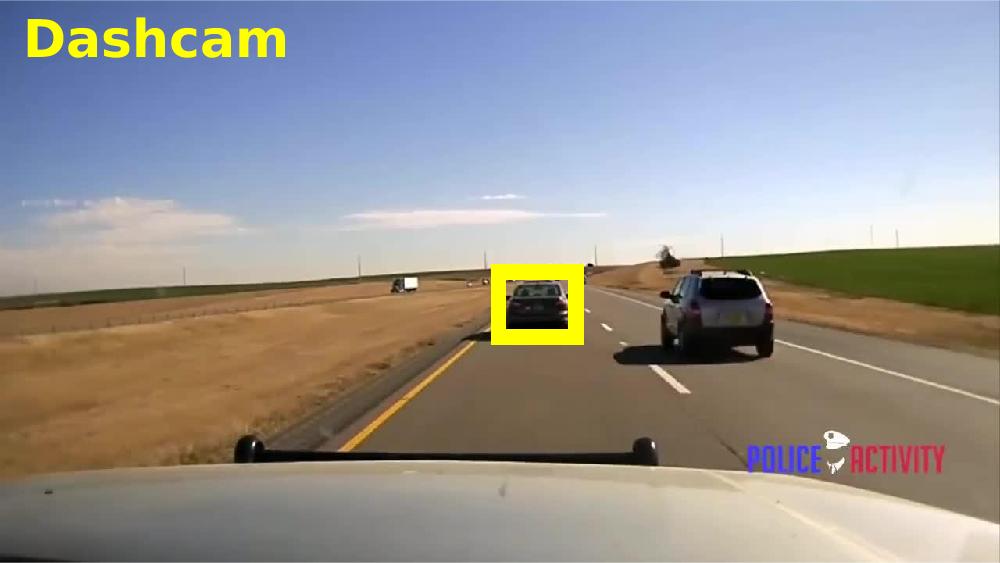}
        \\ 
        \includegraphics[width=0.2\linewidth]{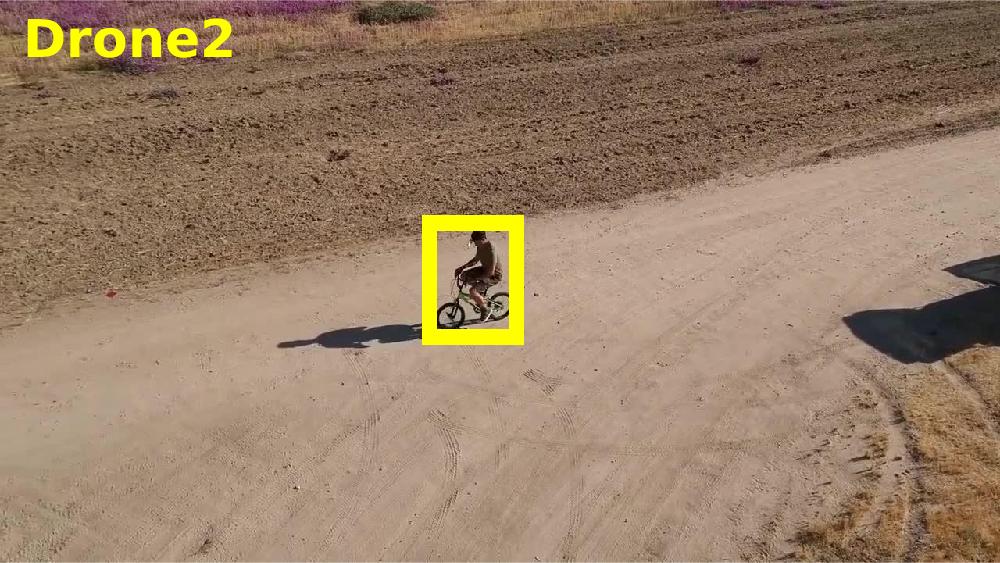}& 
        \includegraphics[width=0.2\linewidth]{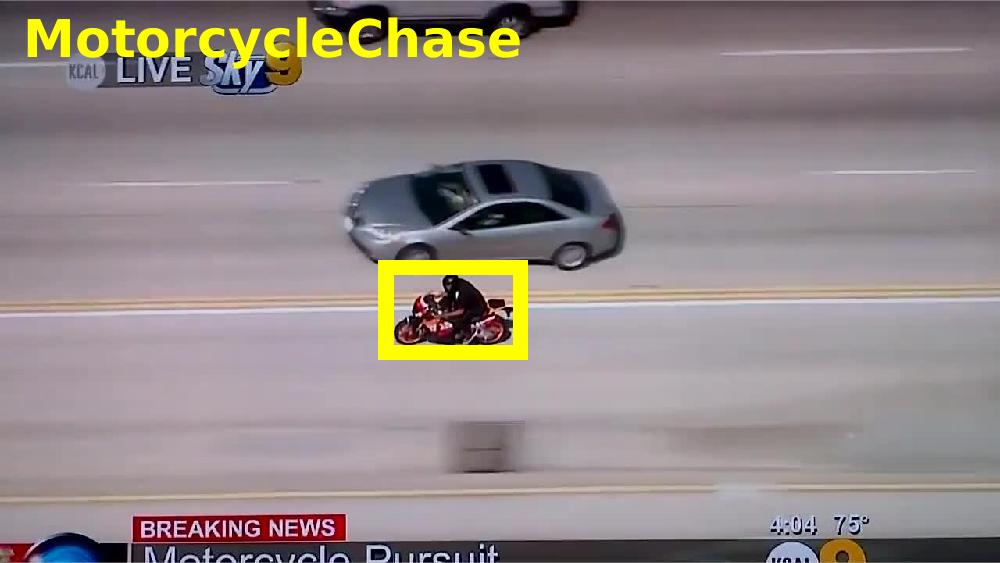}& 
        \includegraphics[width=0.2\linewidth]{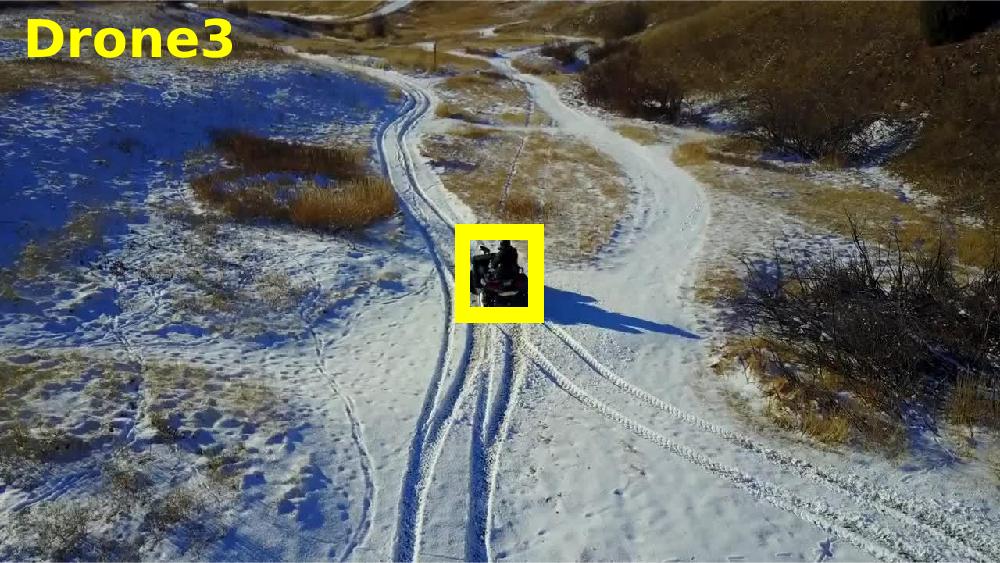}&
        \includegraphics[width=0.2\linewidth]{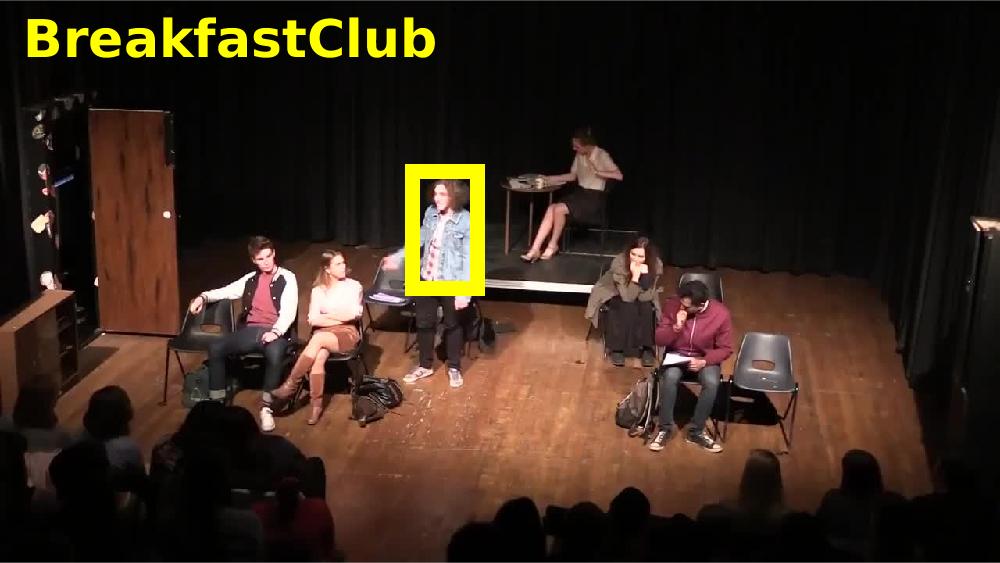}&
        \includegraphics[width=0.2\linewidth]{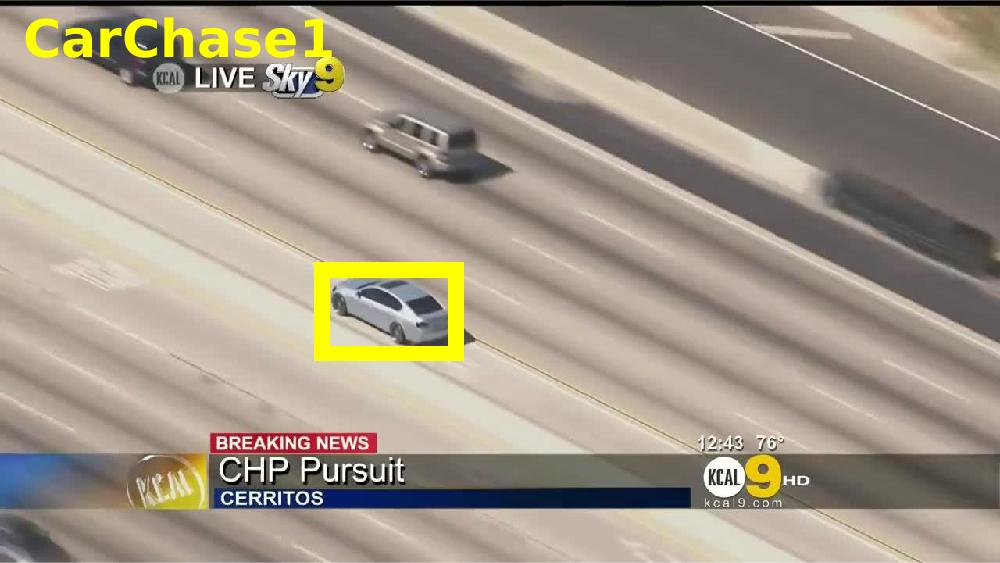}
        \\ 
        \includegraphics[width=0.2\linewidth]{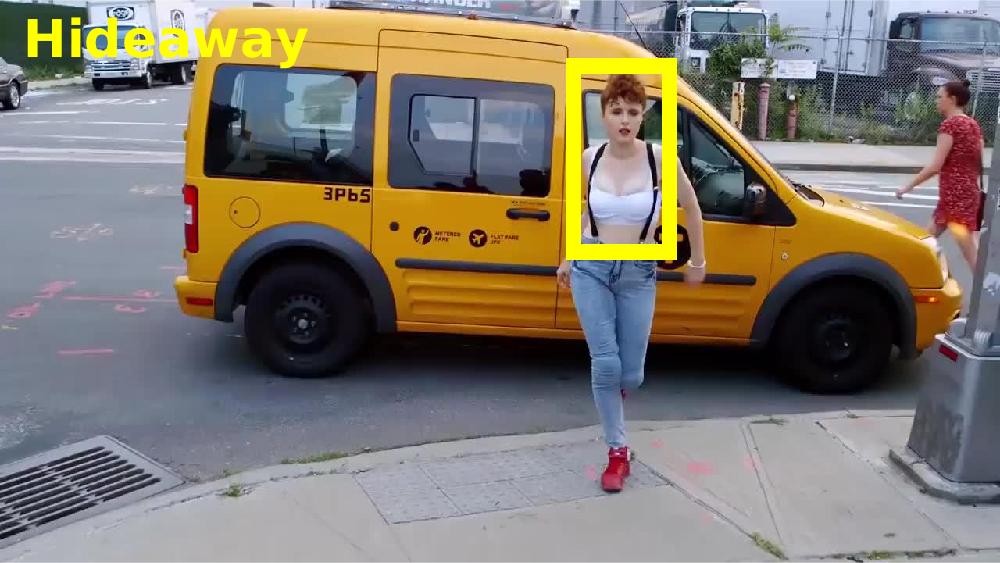}& 
        \includegraphics[width=0.2\linewidth]{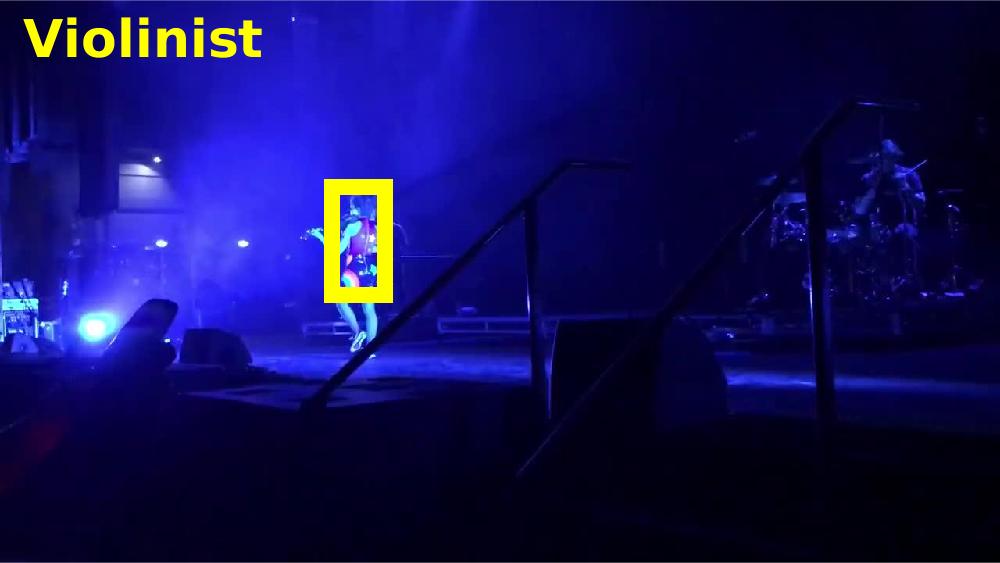}&
        \includegraphics[width=0.2\linewidth]{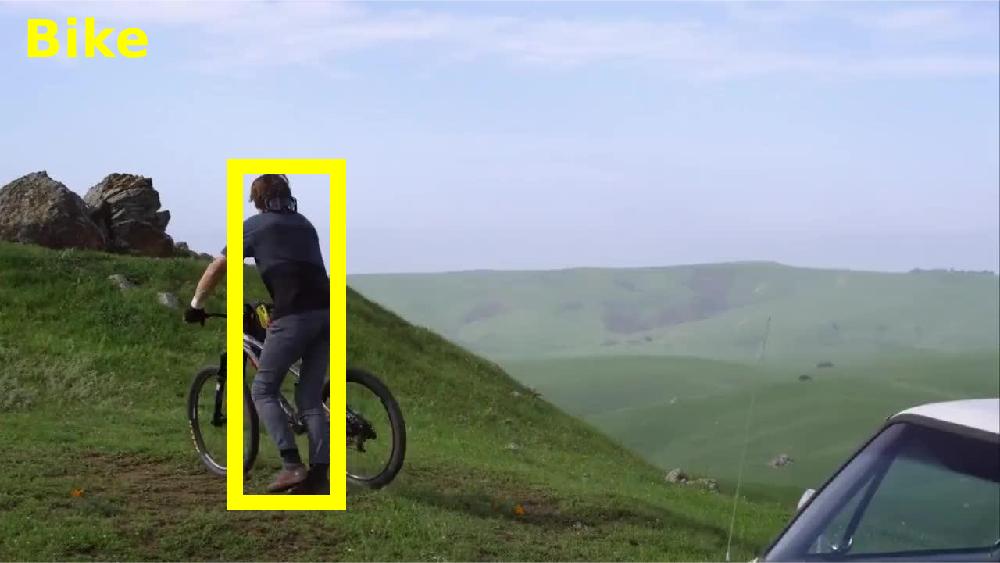}&
        \includegraphics[width=0.2\linewidth]{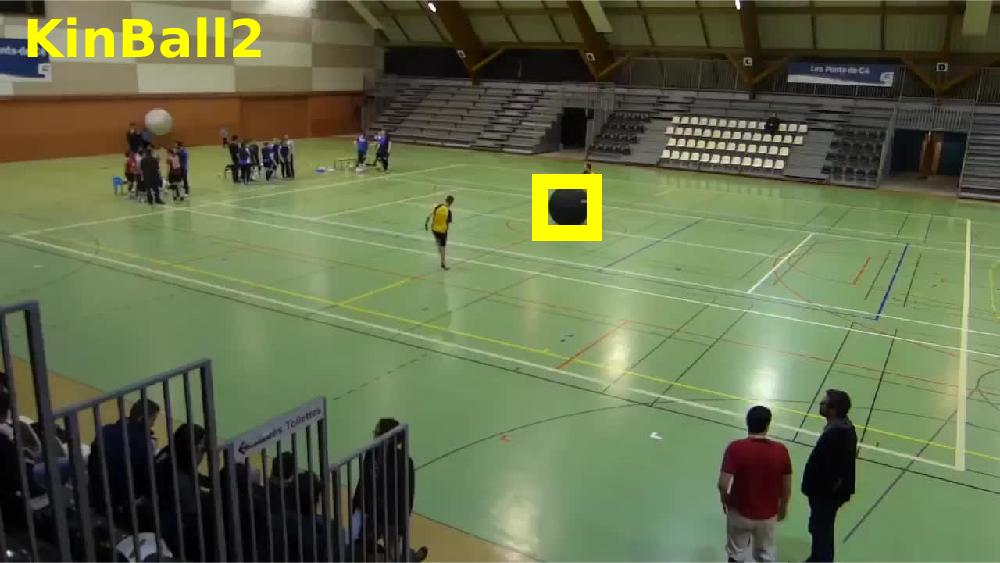}&
        \includegraphics[width=0.2\linewidth]{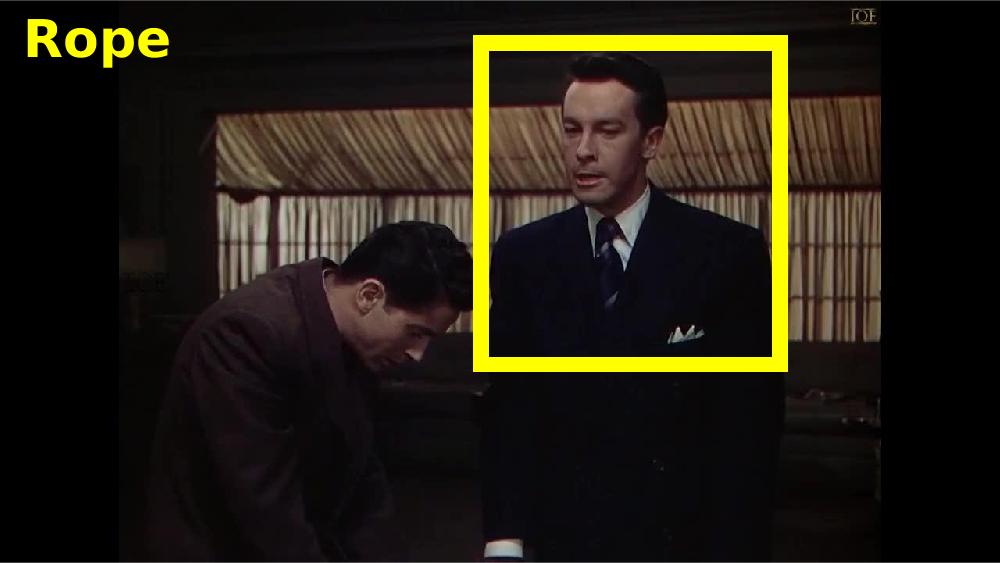} 
        \\ 
        \includegraphics[width=0.2\linewidth]{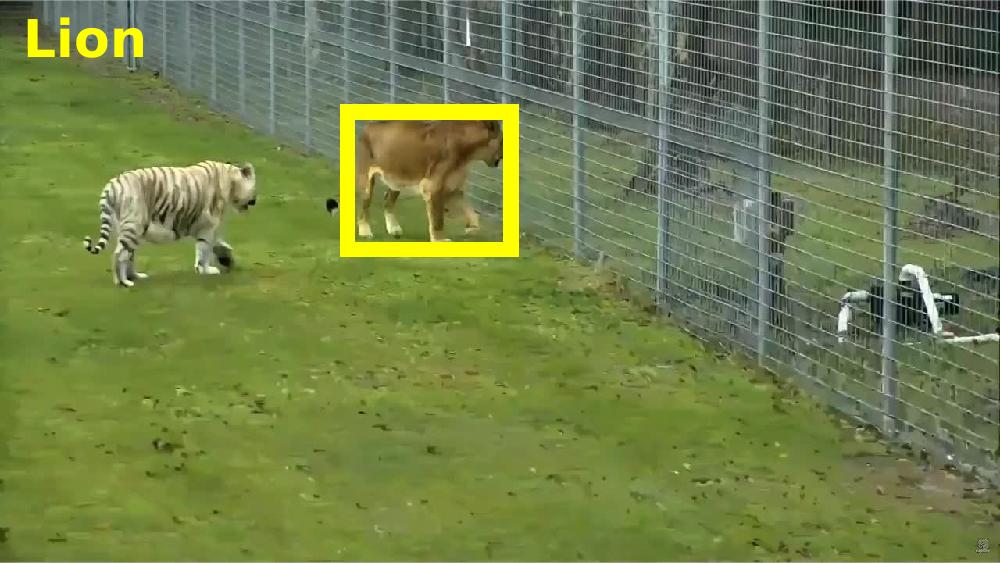}&
        \includegraphics[width=0.2\linewidth]{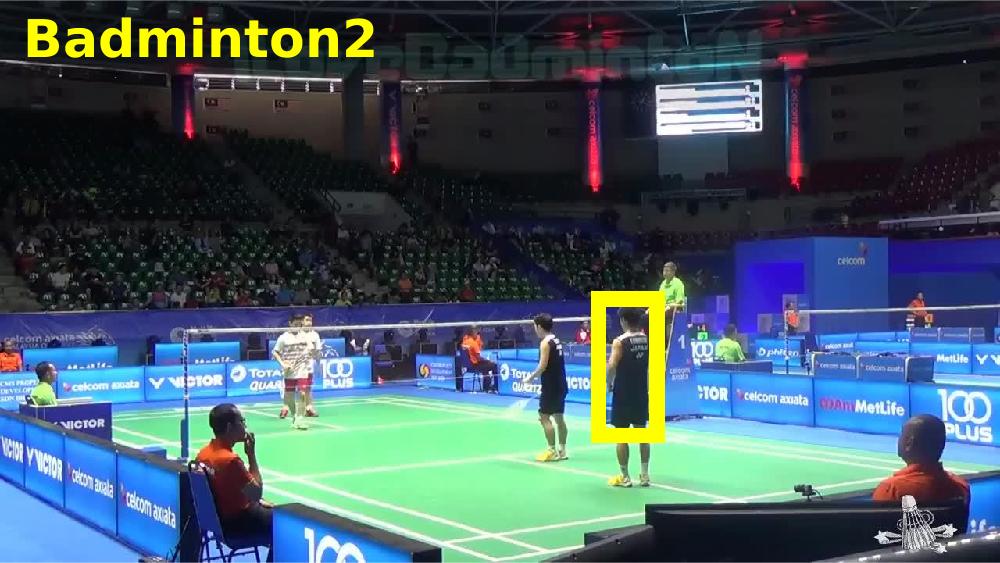}&
        \includegraphics[width=0.2\linewidth]{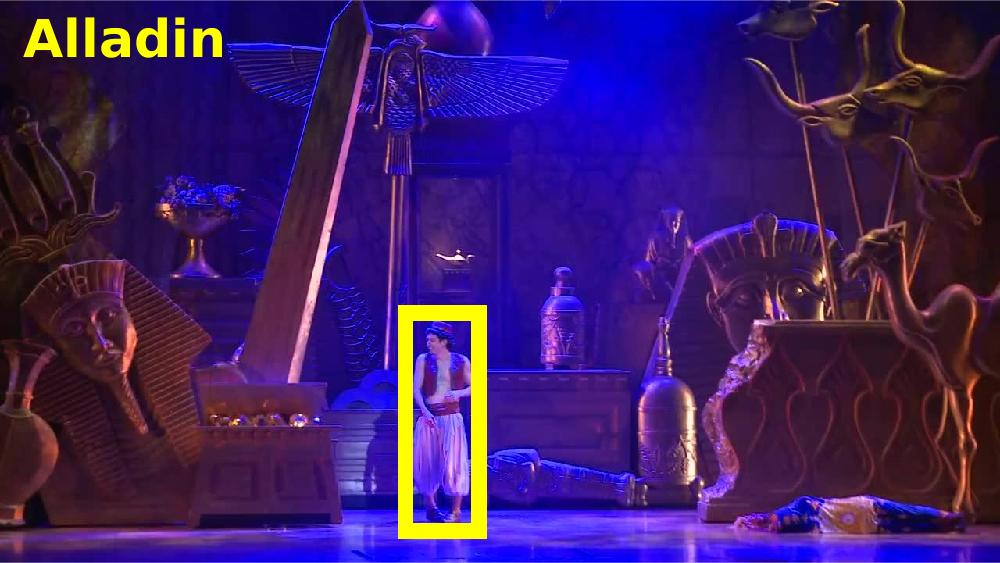}&
        \includegraphics[width=0.2\linewidth]{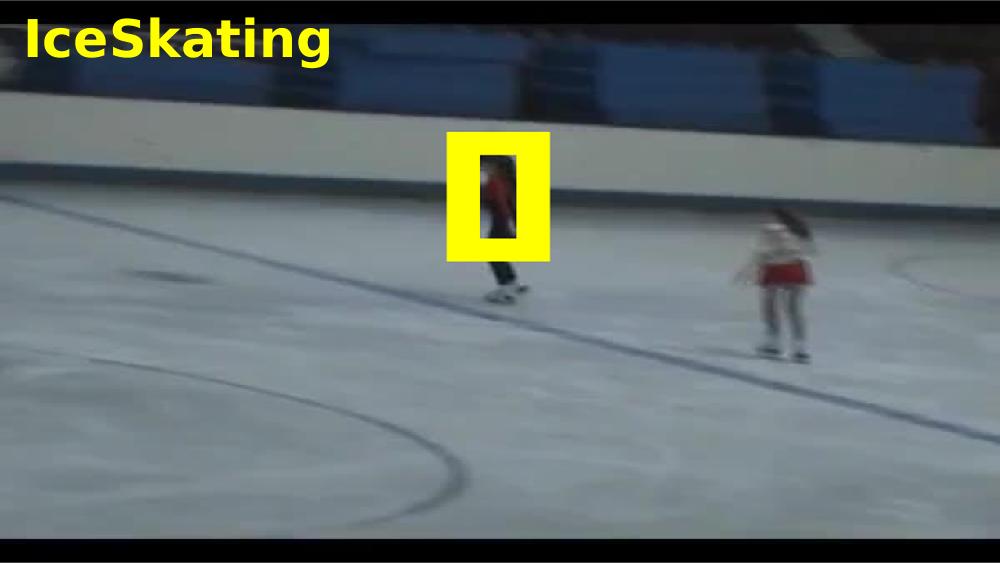}&
        \includegraphics[width=0.2\linewidth]{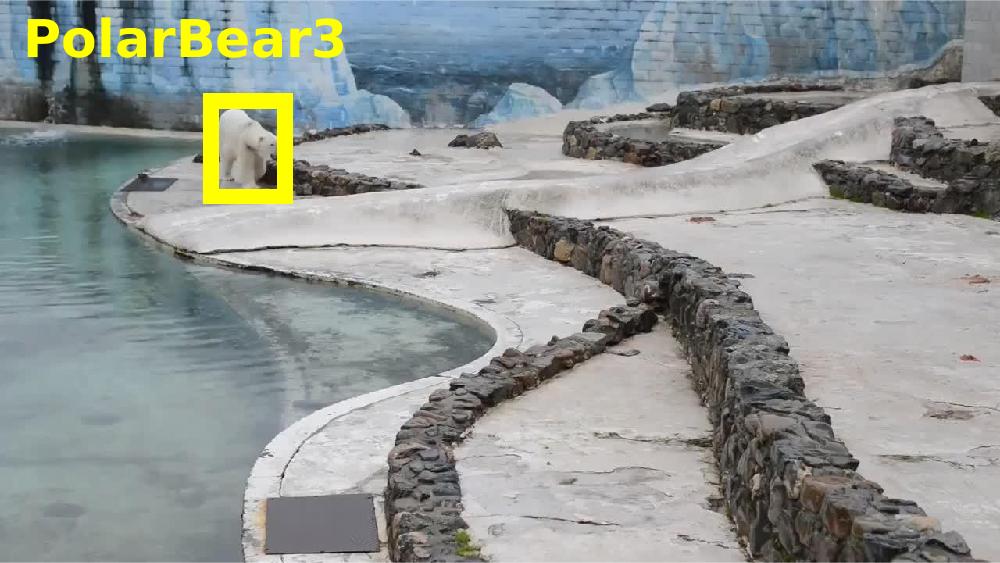}
        \\ 
        \includegraphics[width=0.2\linewidth]{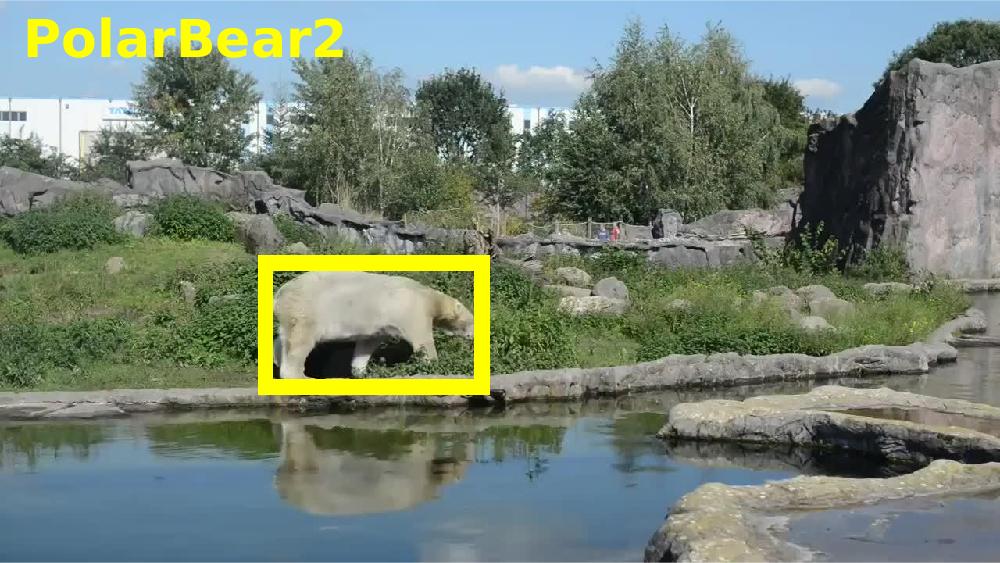}&
        \includegraphics[width=0.2\linewidth]{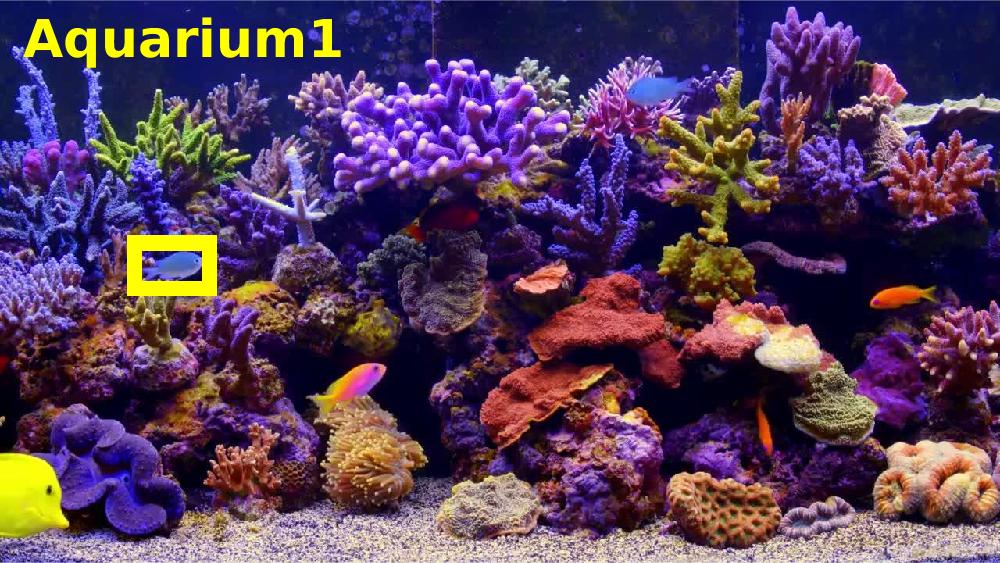}&
        \includegraphics[width=0.2\linewidth]{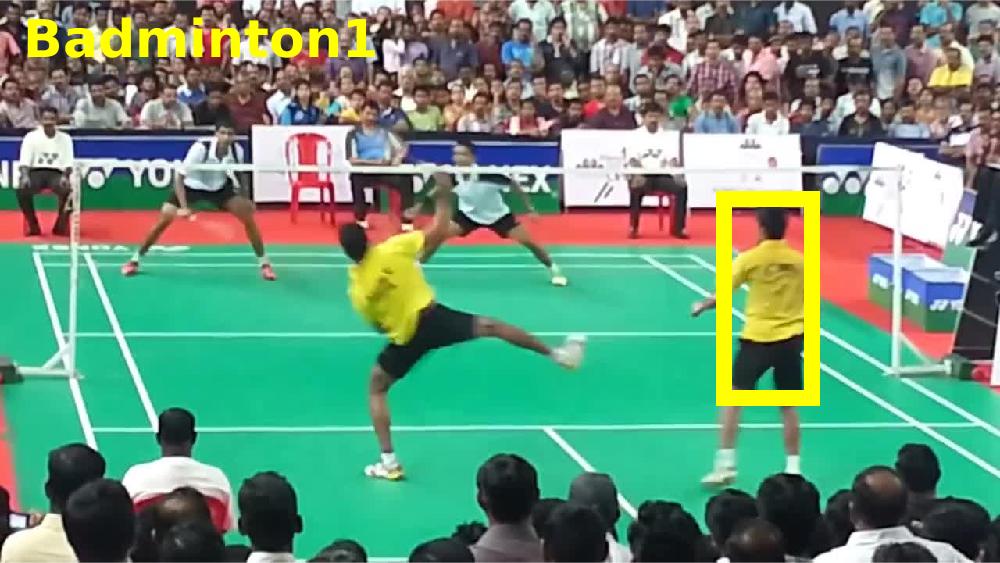}&
        \includegraphics[width=0.2\linewidth]{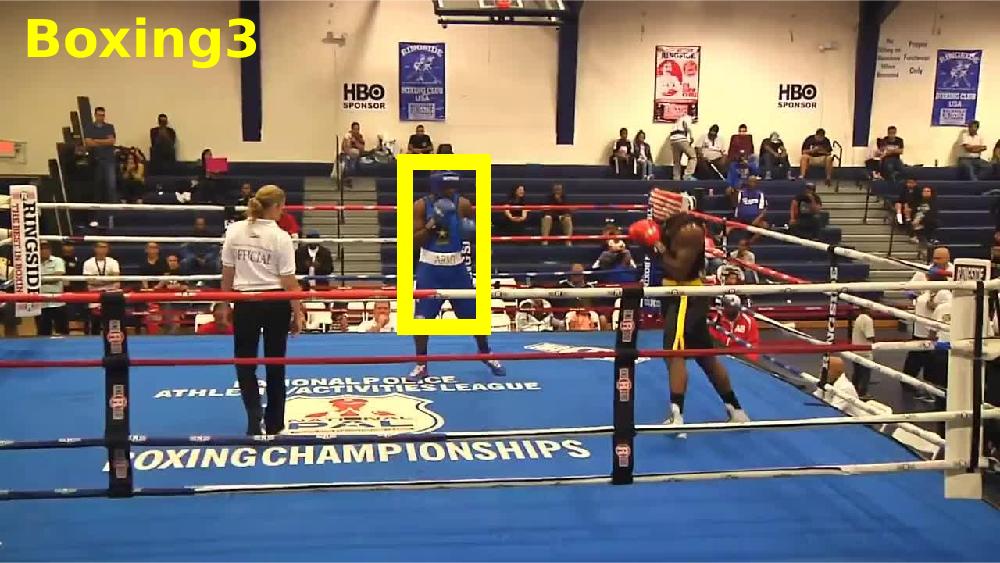}&
        \includegraphics[width=0.2\linewidth]{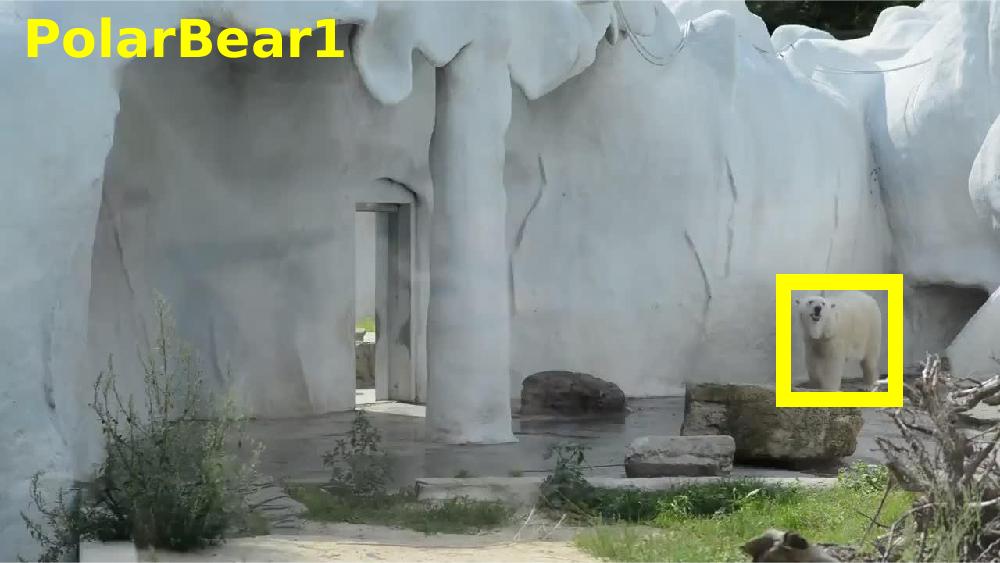}
        \\ 
        \includegraphics[width=0.2\linewidth]{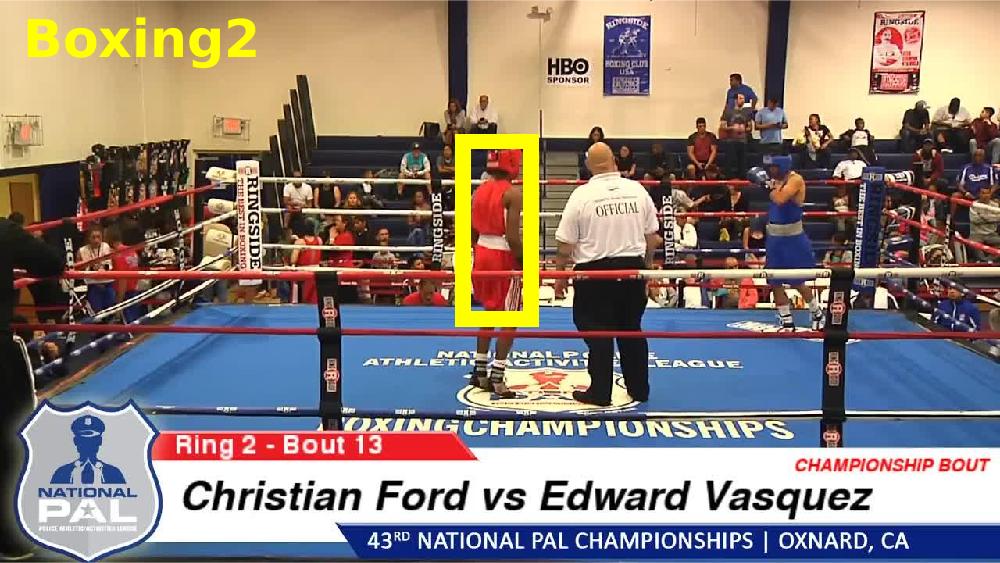}&
        \includegraphics[width=0.2\linewidth]{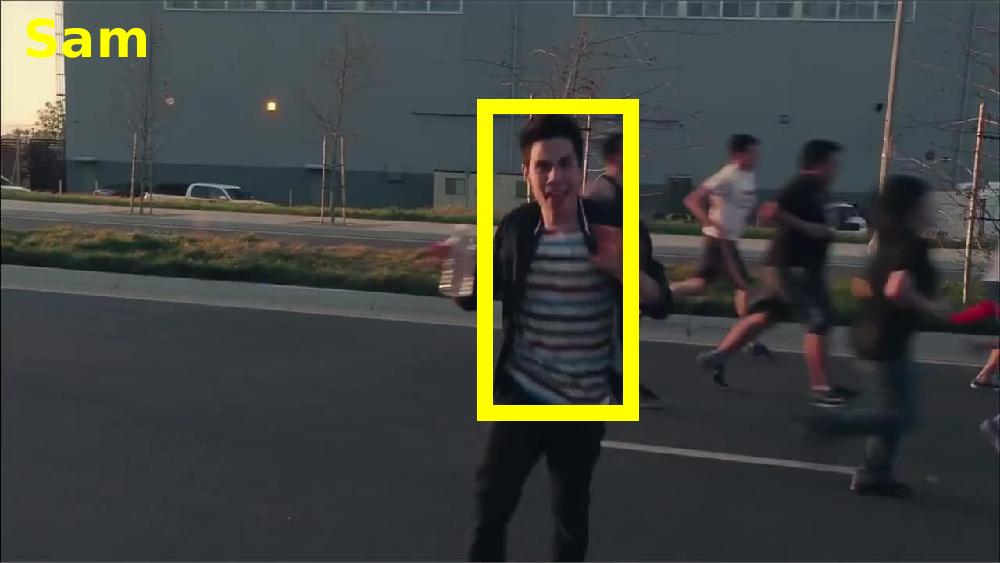}&
        \includegraphics[width=0.2\linewidth]{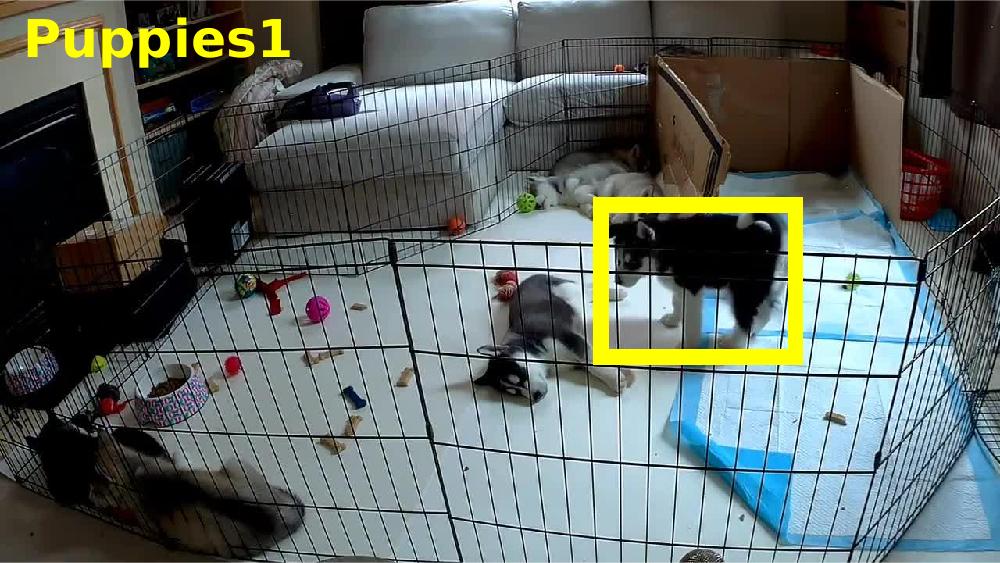}&
        \includegraphics[width=0.2\linewidth]{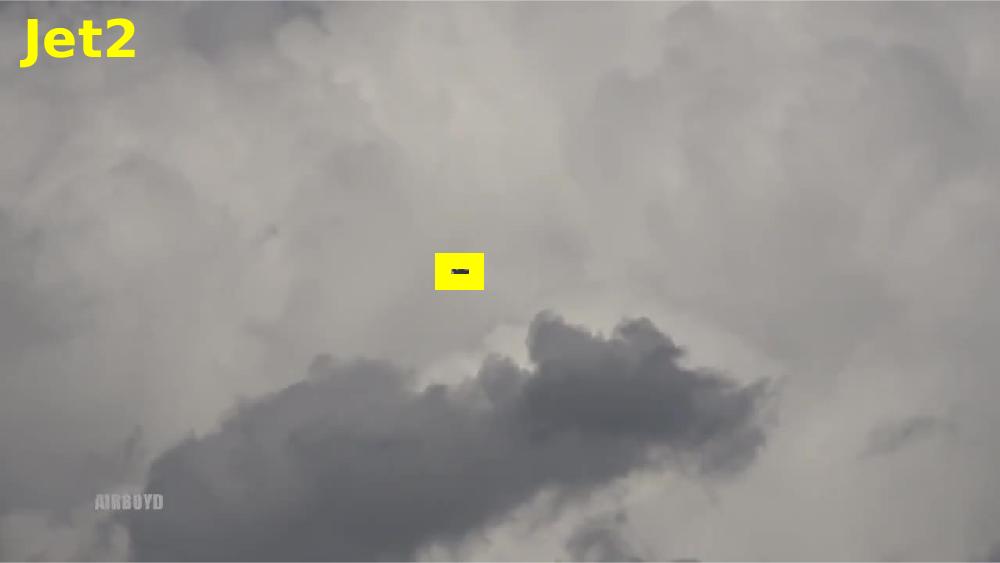}&
        \includegraphics[width=0.2\linewidth]{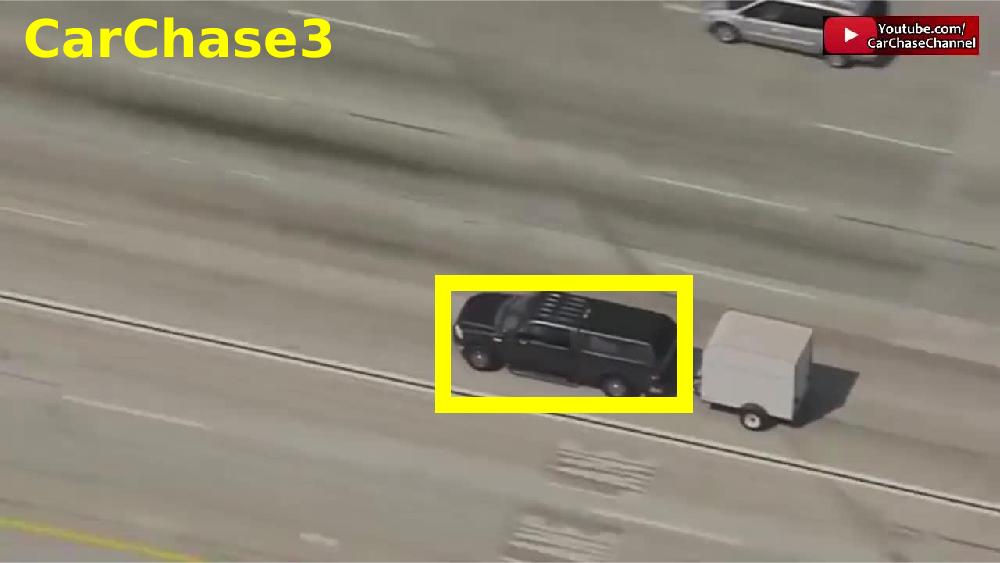} 
        \\ 
        \includegraphics[width=0.2\linewidth]{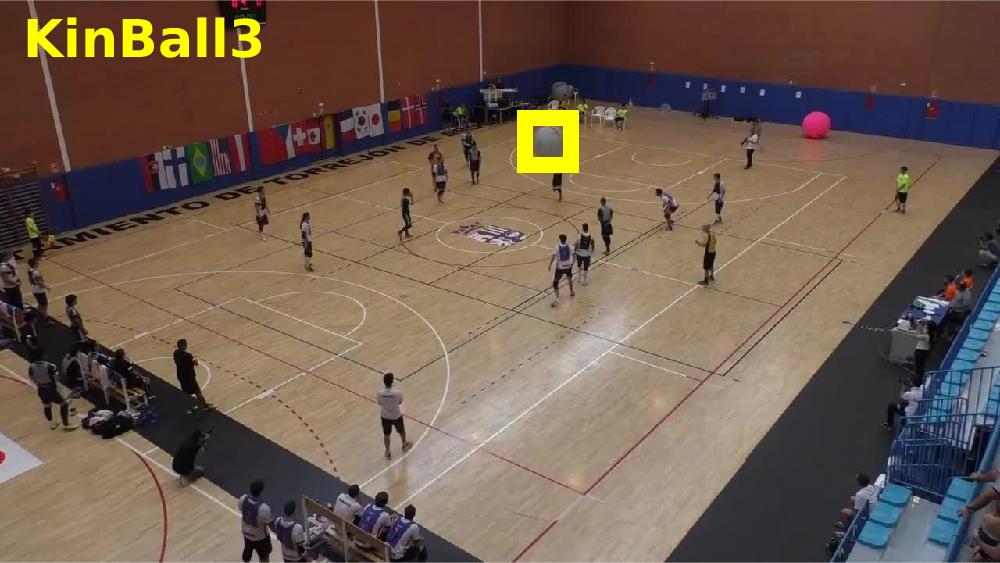}&
        \includegraphics[width=0.2\linewidth]{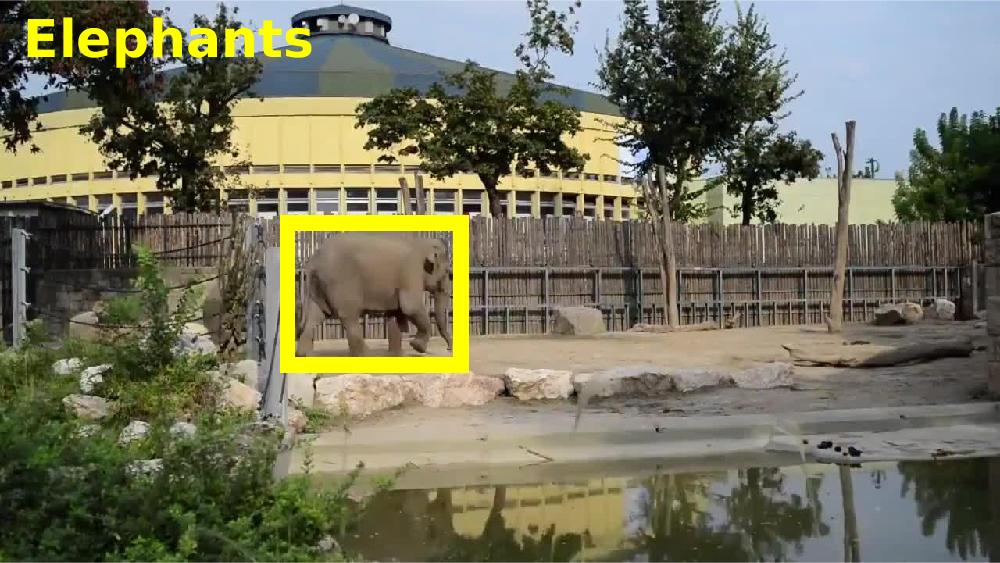}&
        \includegraphics[width=0.2\linewidth]{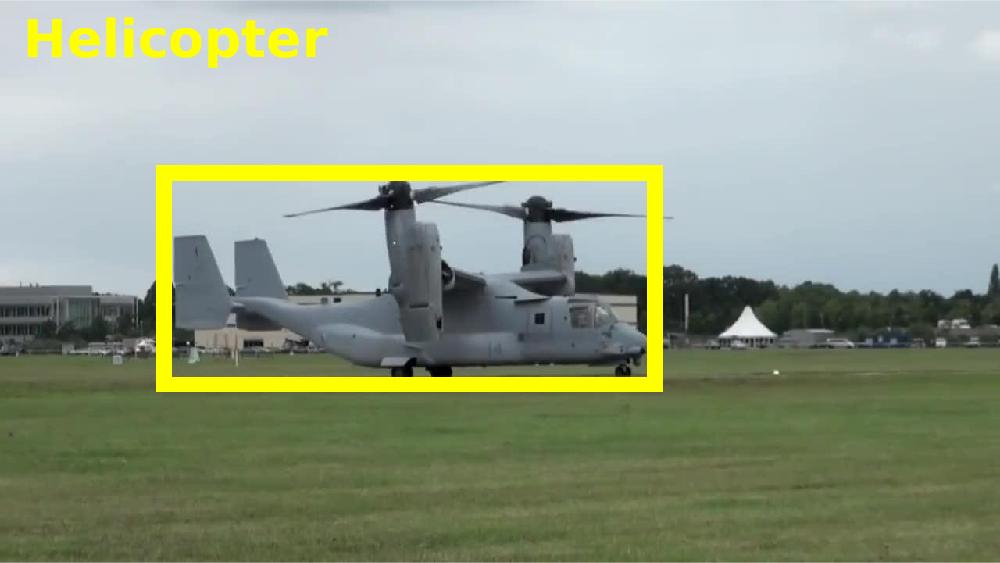}&
        \includegraphics[width=0.2\linewidth]{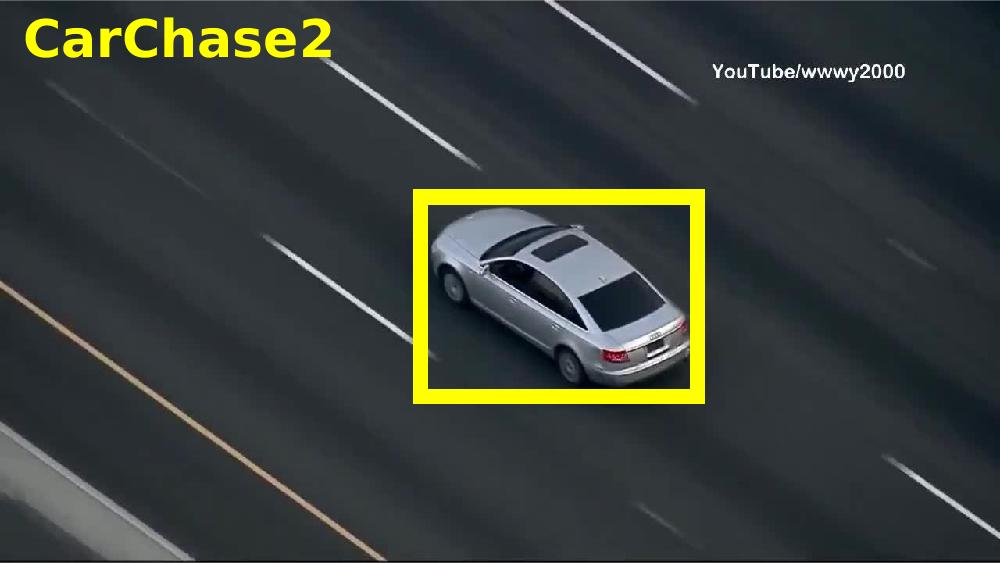}&
        \includegraphics[width=0.2\linewidth]{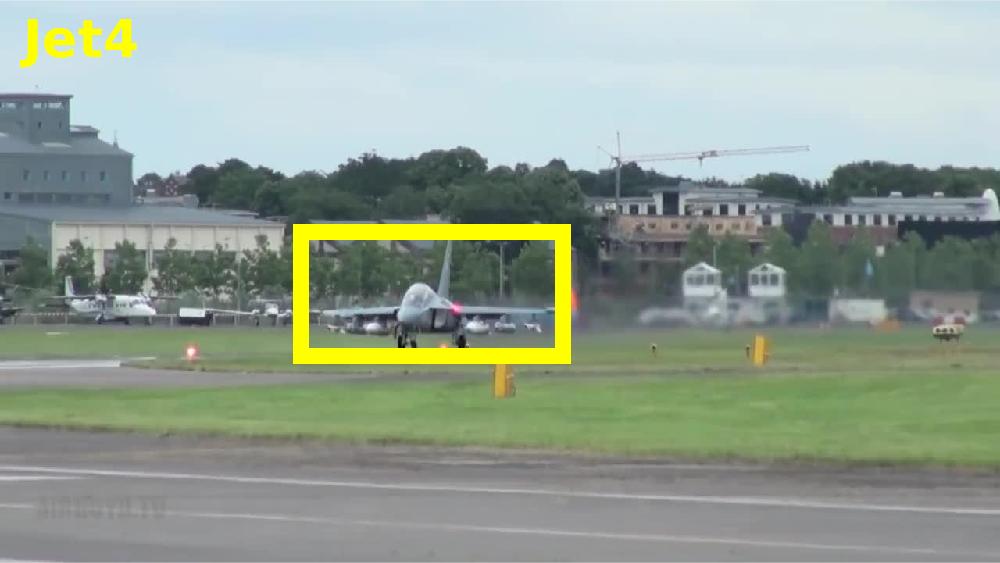}              
        \\ 
        \includegraphics[width=0.2\linewidth]{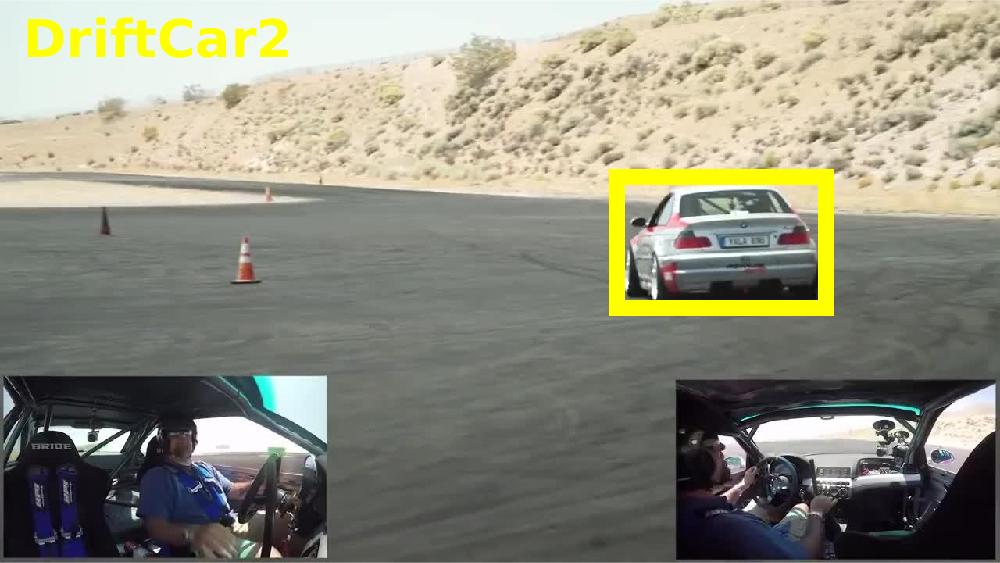}& 
        \includegraphics[width=0.2\linewidth]{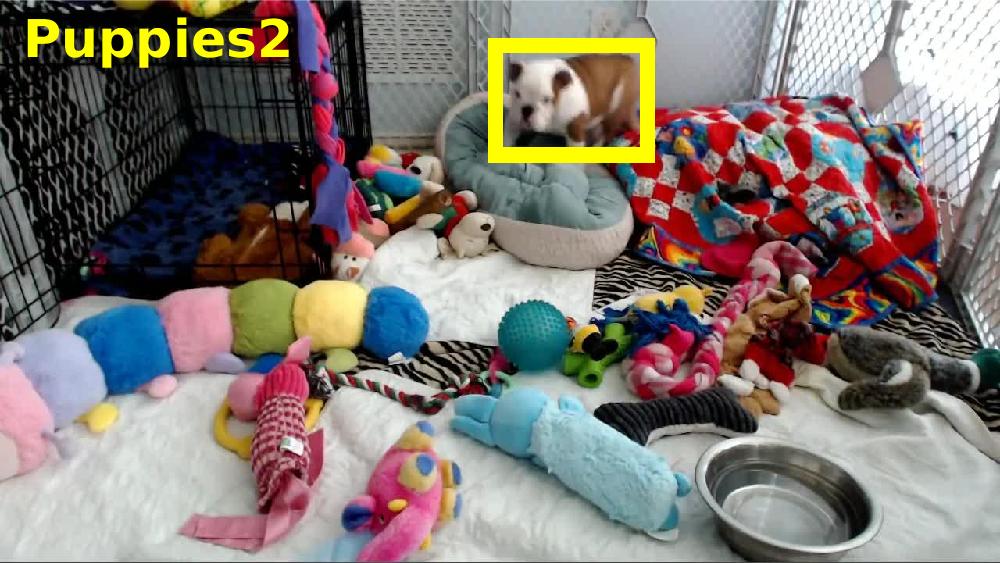}&
        \includegraphics[width=0.2\linewidth]{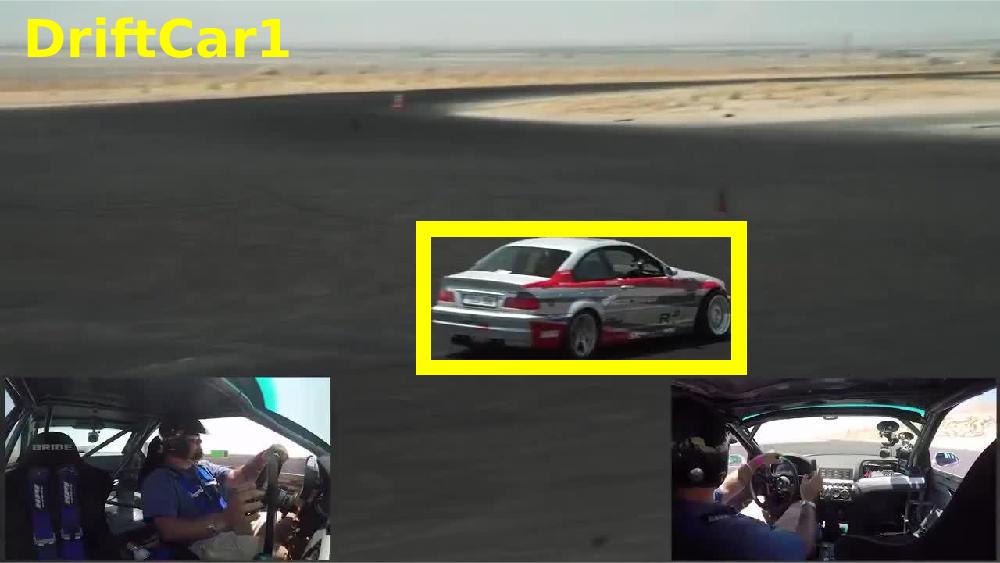}&
        \includegraphics[width=0.2\linewidth]{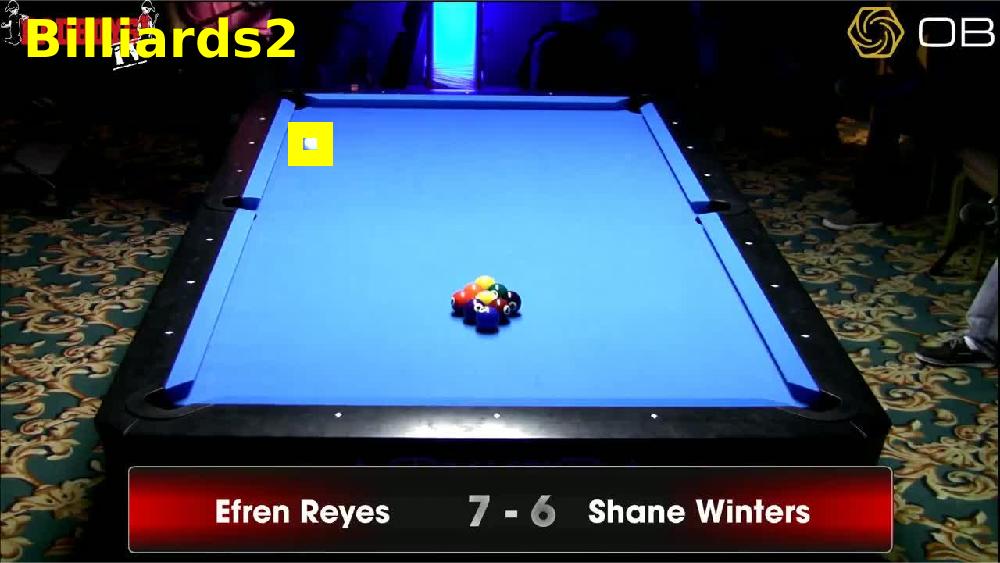} &
        \includegraphics[width=0.2\linewidth]{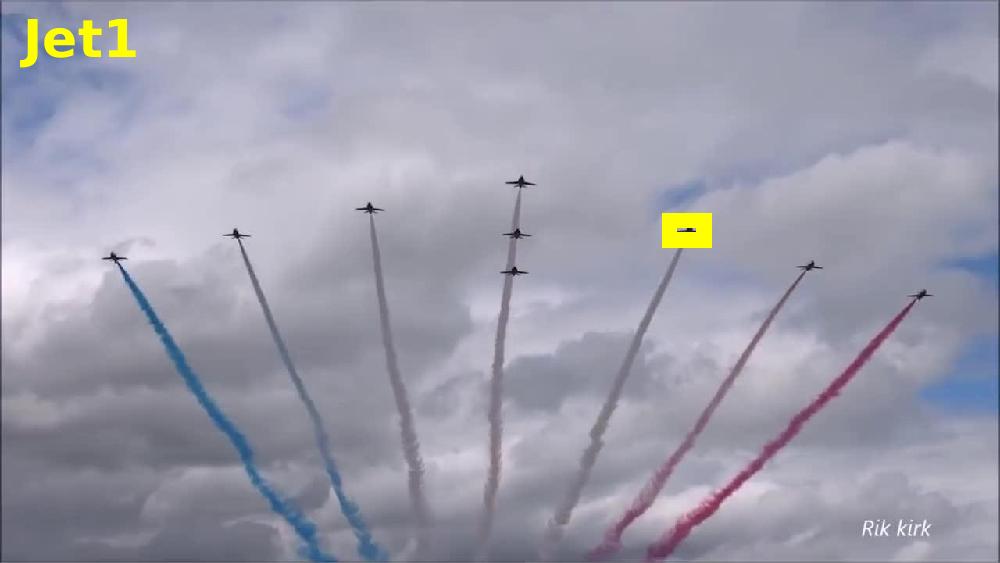}
        \\ 
        \includegraphics[width=0.2\linewidth]{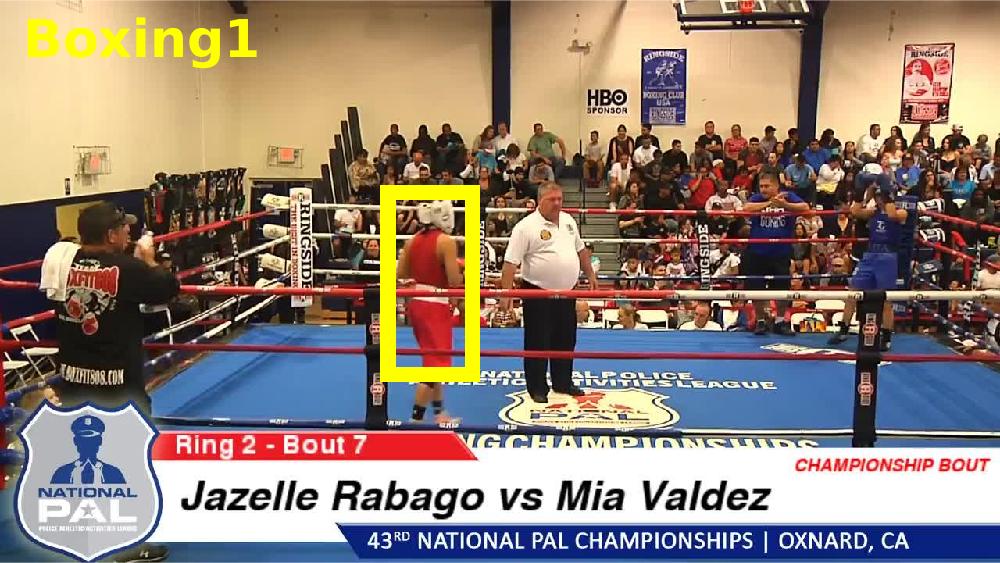}&
        \includegraphics[width=0.2\linewidth]{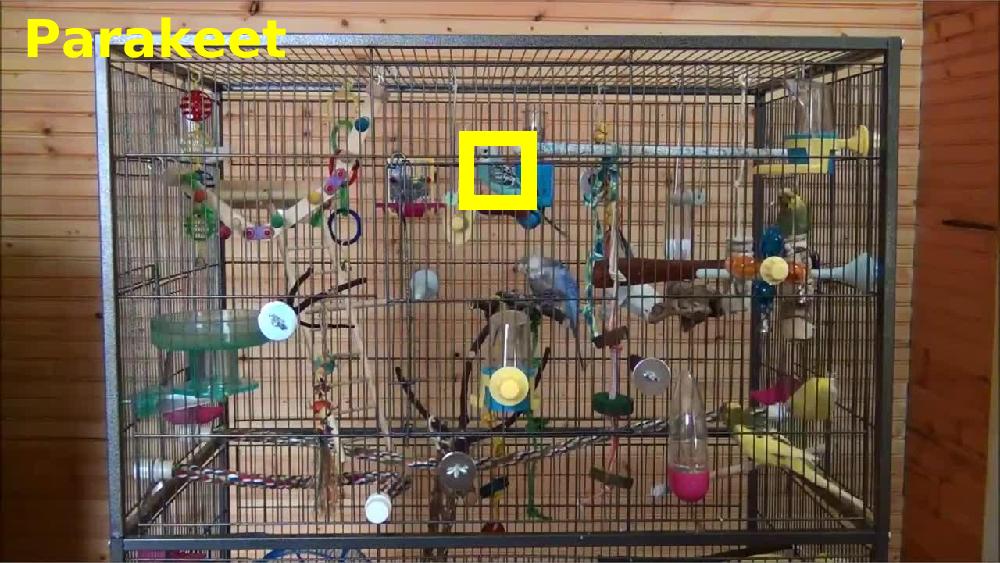}&
        \includegraphics[width=0.2\linewidth]{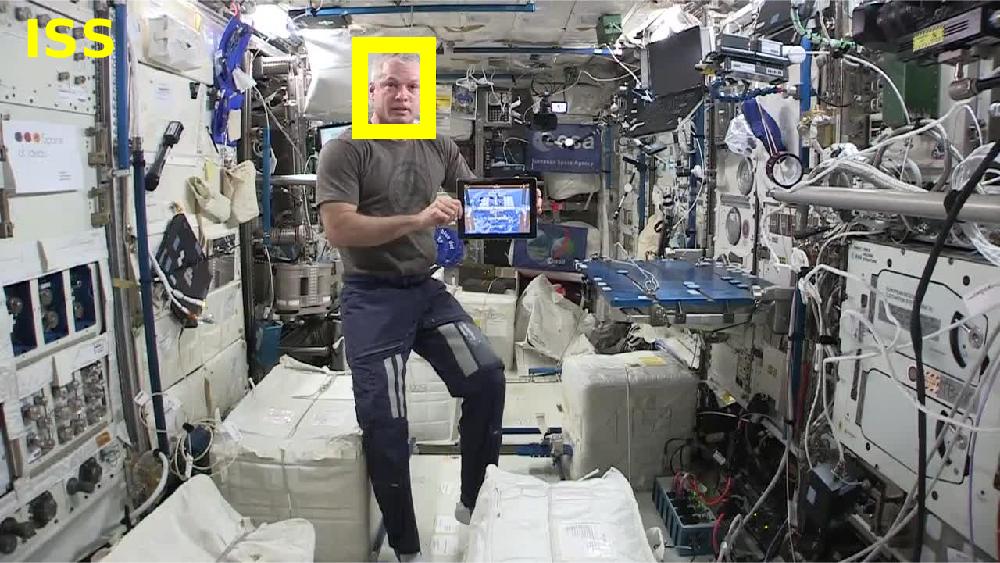}&
        \includegraphics[width=0.2\linewidth]{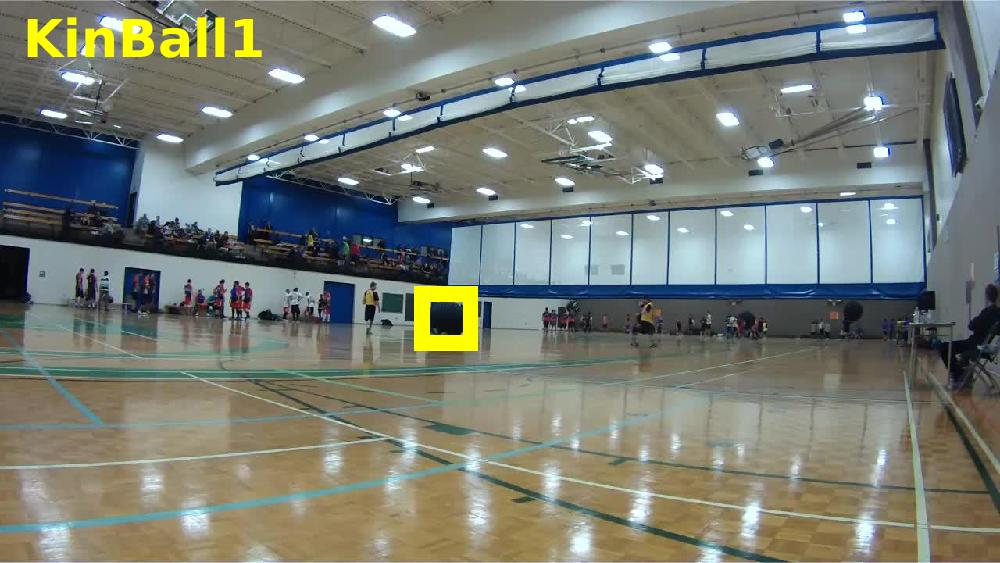}&
        \includegraphics[width=0.2\linewidth]{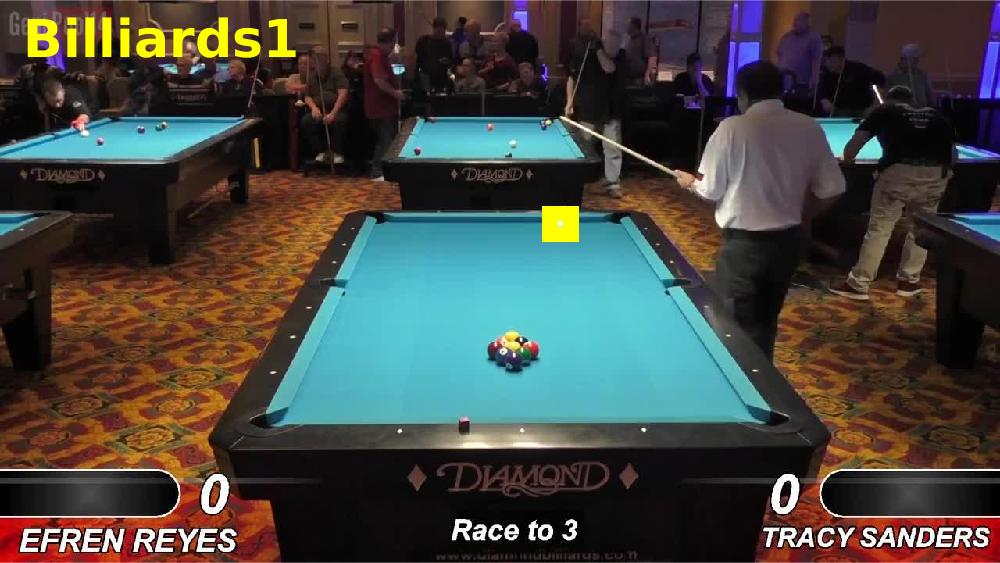}
        \\
        \includegraphics[width=0.2\linewidth]{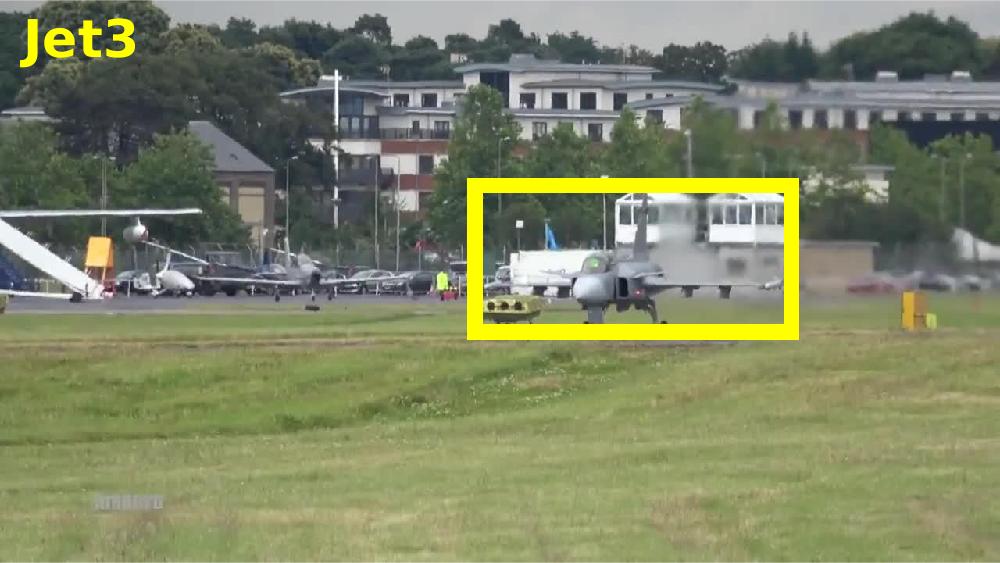}&
        \includegraphics[width=0.2\linewidth]{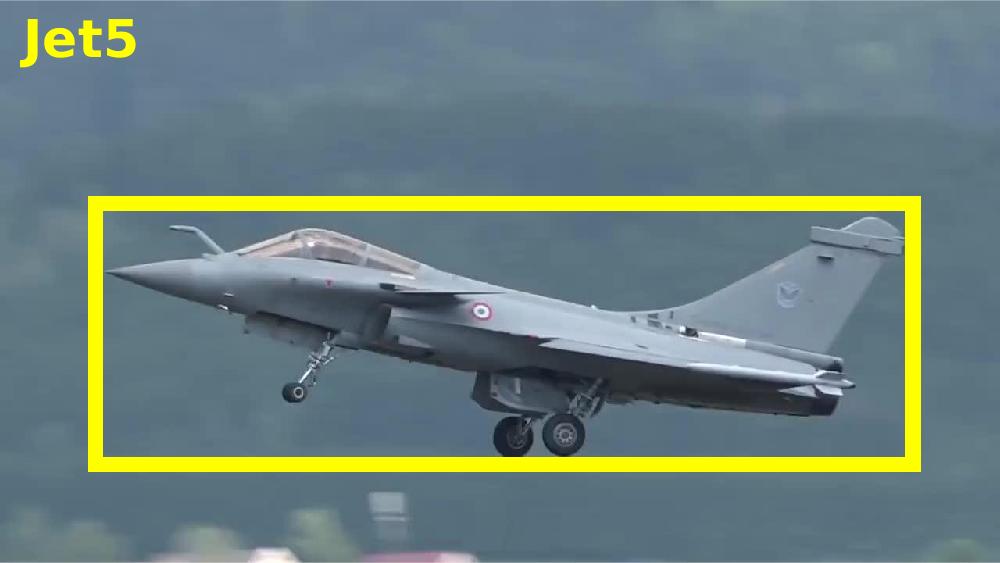}&
        \includegraphics[width=0.2\linewidth]{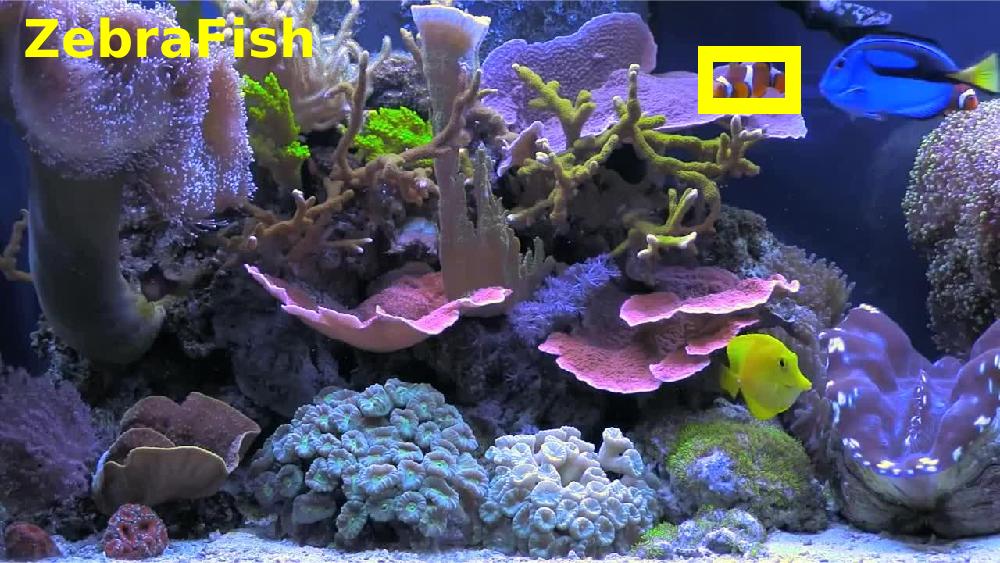}&        
        \includegraphics[width=0.2\linewidth]{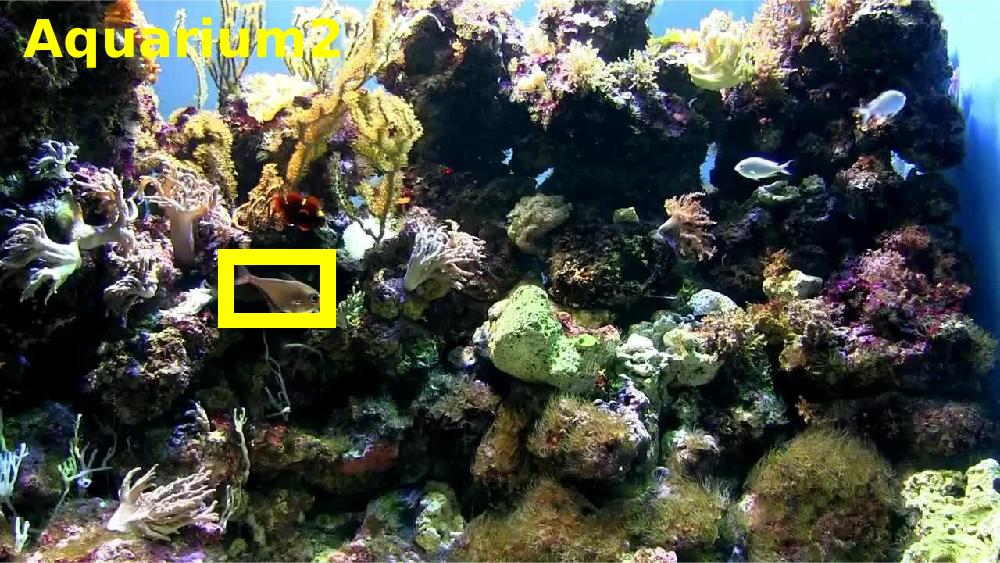}&
        \includegraphics[width=0.2\linewidth]{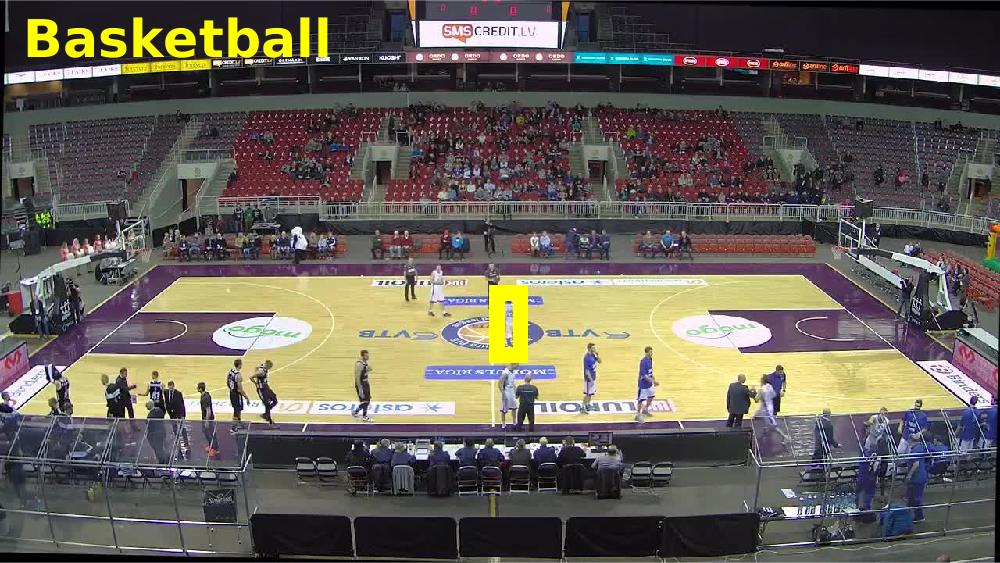}    
        \end{tabular}
        \caption{First frames of all the 50 sequences of TLP dataset. The sequences are sorted in ascending order on the basis of mean success rate (defined in Section \ref{section:eval}) of all trackers at IoU threshold of 0.5. The sequences at the bottom right are more difficult to track than the ones at the top left.}
        \label{fig:first_frames}
\end{figure}

Deep learning based trackers present another paradigm in visual object tracking. Data deficiency appeared to be the major limitation in early attempts~\cite{wang2013learning,li2014robust}. Later approaches~\cite{wang2015transferring,hong2015online} learnt offline tasks like objectness or saliency offline from object detection datasets and benefited from it during online tracking. However, the gap between the two tasks turned out to be a limitation in these works. The work by Nam \etal~\cite{nam2016mdnet} proposed a novel direction by posing the tracking problem as evaluating the positive and negative candidate windows randomly sampled around the previous target state. They proposed a two phase training, first domain dependent and offline to fix initial layers and second online phase to update the last fully connected layer. 

Siamese frameworks~\cite{bertinetto2016fully,held2016learning} have been proposed to regress the location of the target, given the previous frame location. They either employ data augmentation using affine transformations on individual images~\cite{held2016learning} or use video detection datasets for large scale offline training~\cite{bertinetto2016fully}. In both these approaches, the network is evaluated without any fine tuning at the test time, which significantly increases their computational efficiency. However, this comes at the cost of losing ability to update appearance models or learn target specific information, which may be crucial for visual tracking. More recently, Yun \etal~\cite{yun2017adnet} proposed a tracker controlled by action-decision network (ADNet), which pursues the target object by sequential actions iteratively, utilizing reinforcement learning for visual tracking.  

\section{TLP Dataset} \label{section:dataset}
\begin{figure}[t]
        \centering
        \begin{tabular}[b]{c c c}
        \includegraphics[width=0.32\linewidth]{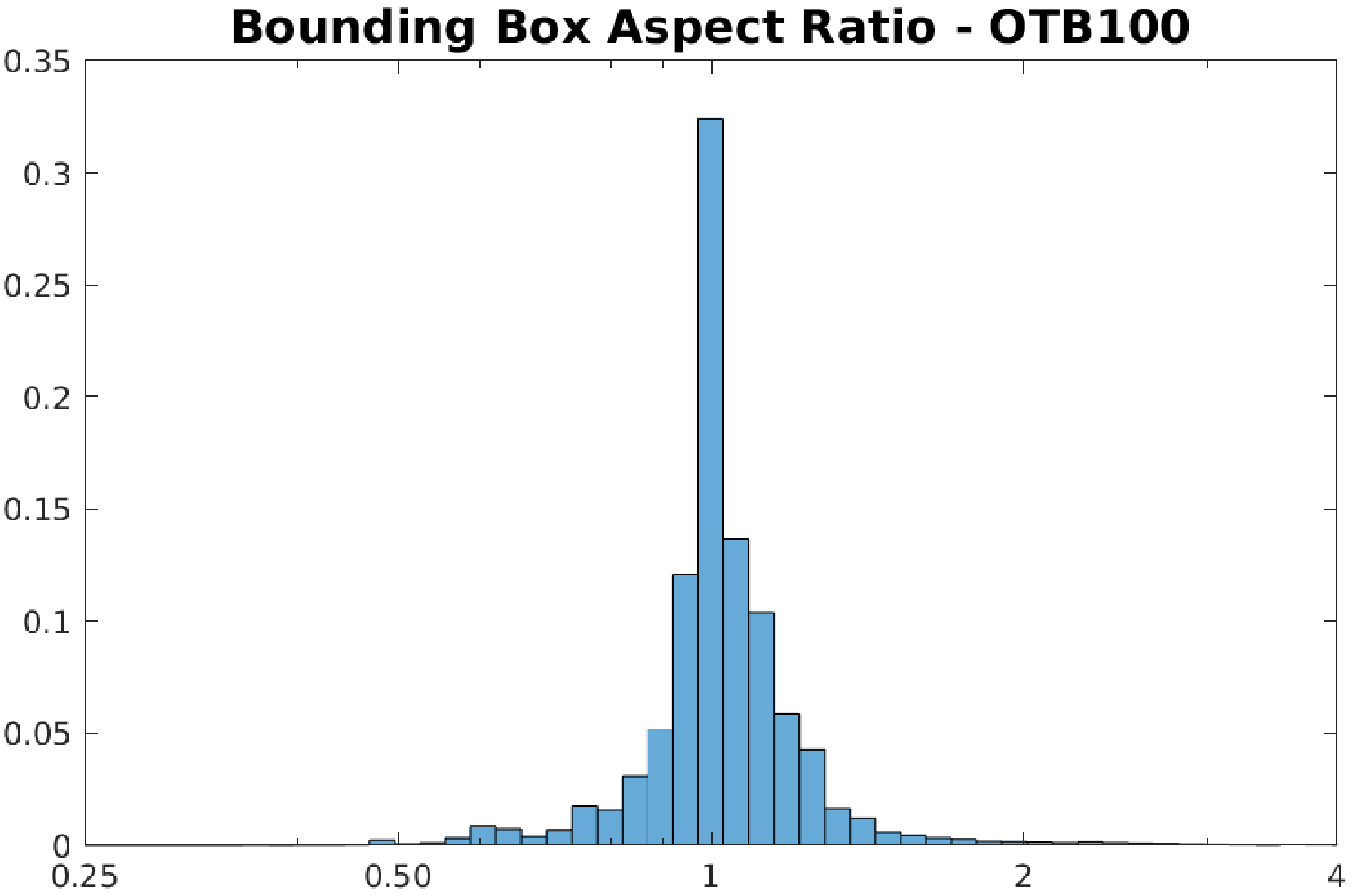}& 
        \includegraphics[width=0.32\linewidth]{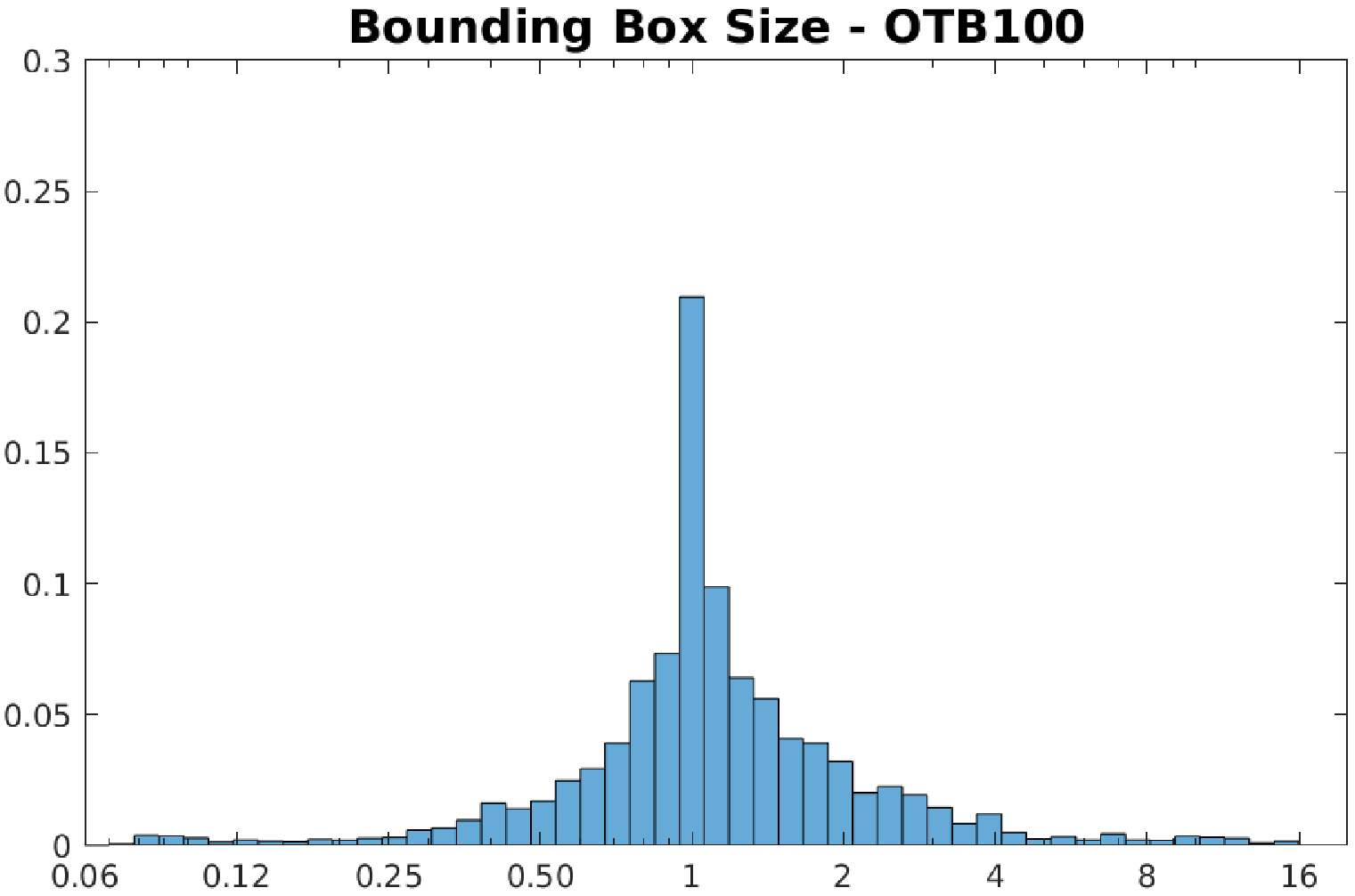}& 
        \includegraphics[width=0.32\linewidth]{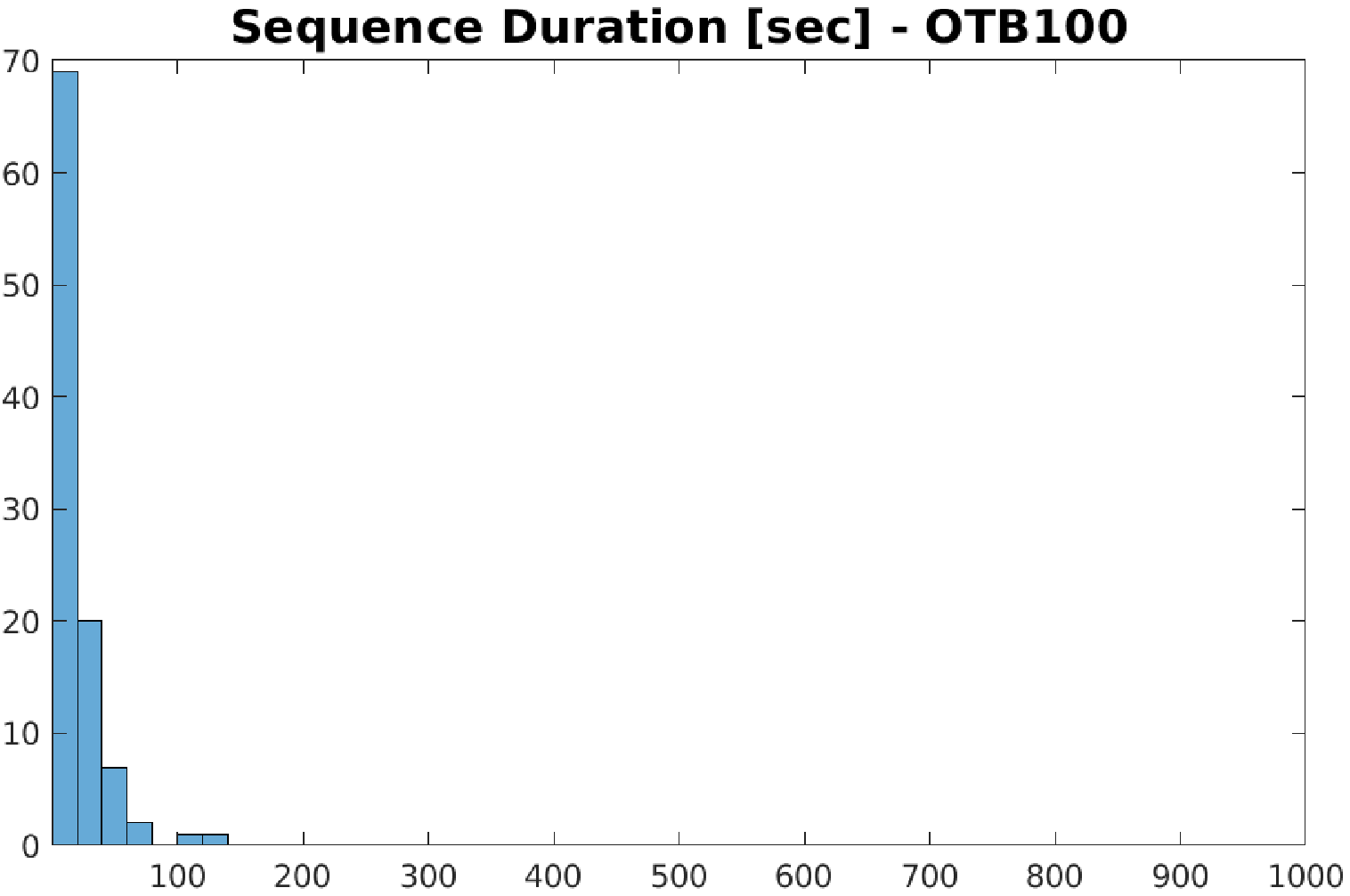} \\
        \includegraphics[width=0.32\linewidth]{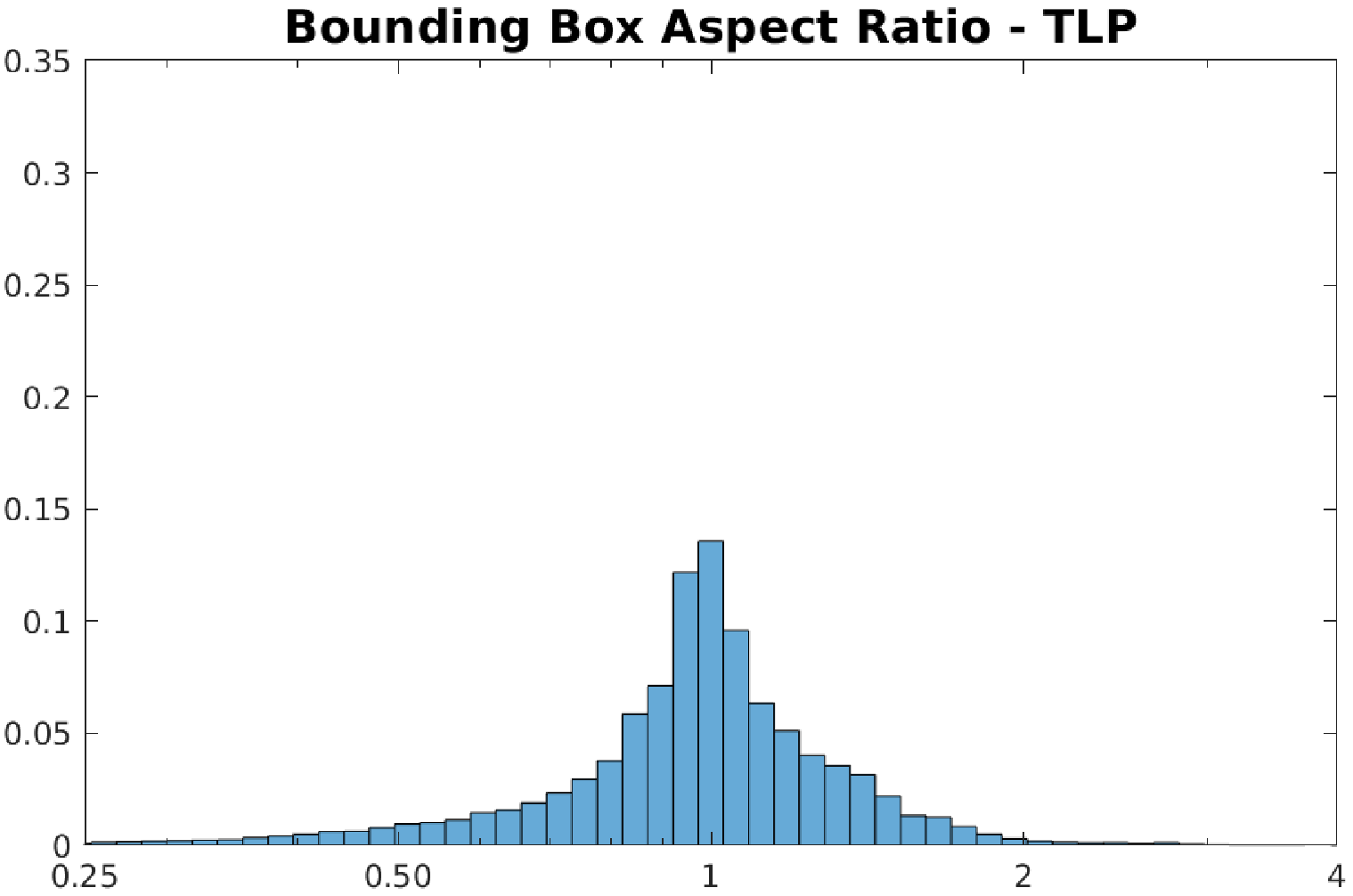}& 
        \includegraphics[width=0.32\linewidth]{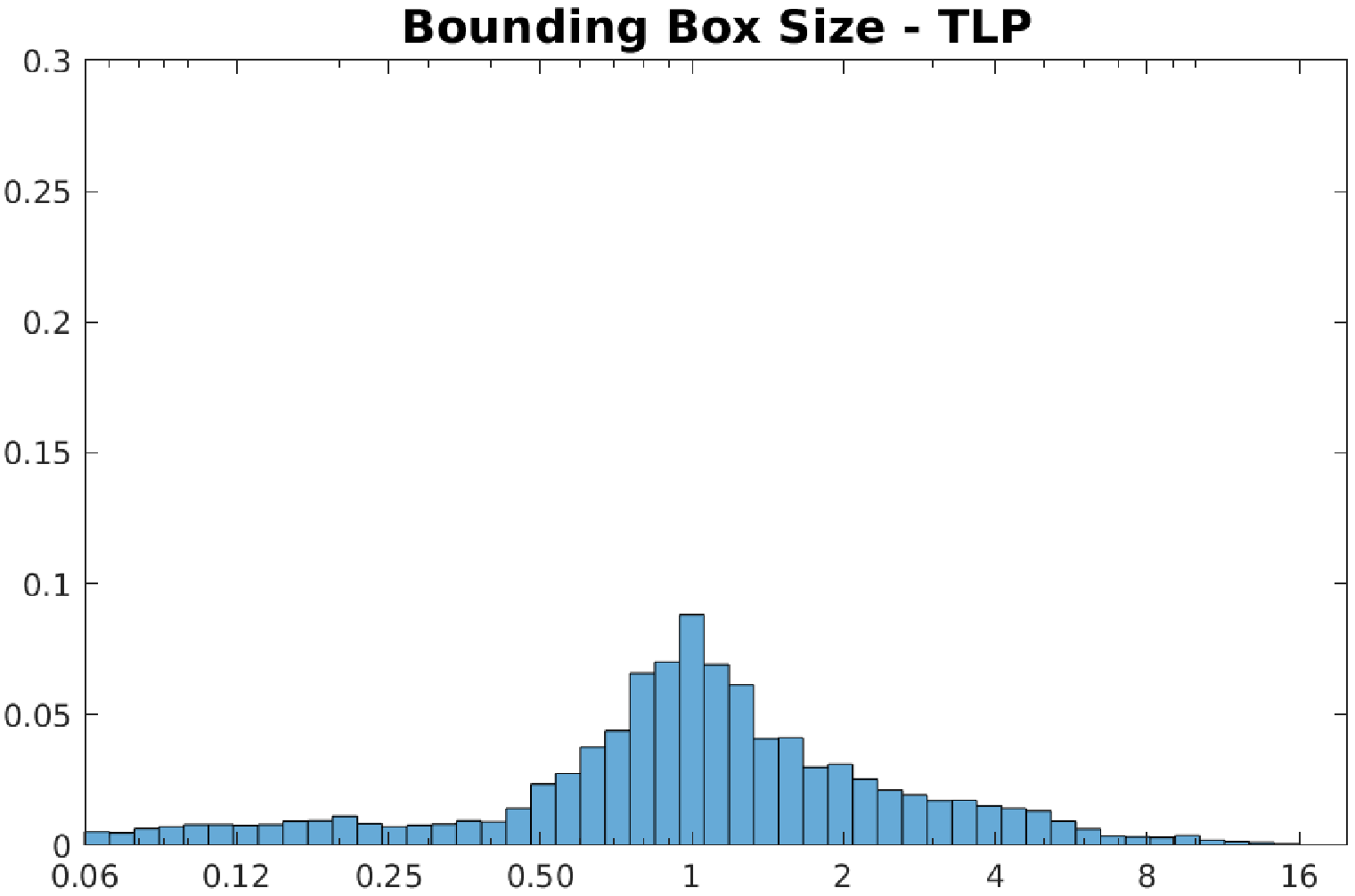}& 
        \includegraphics[width=0.32\linewidth]{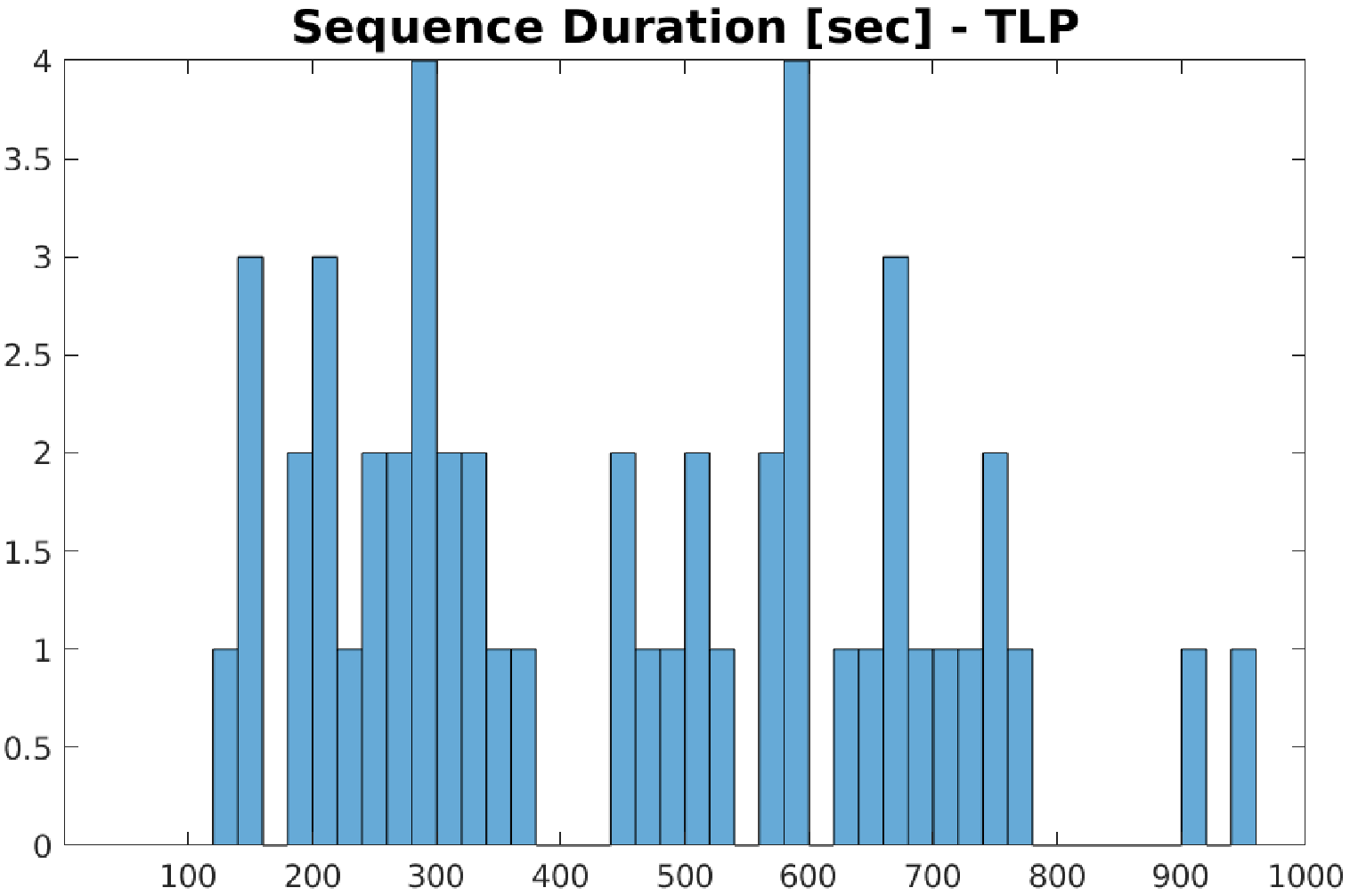} \\  
        \end{tabular}
        \caption{Column 1 and 2: Proportional change of the targets aspect ratio and bounding box size (area in pixels) with respect to the first frame in OTB100 and TLP. Results are compiled over all sequences in each dataset as a histogram with log scale on the x-axis. Column 3: Histogram of sequence duration (in seconds) across the two datasets.}
        \label{fig:compare_OTB}
\end{figure}

The TLP dataset consists of 50 videos collected from YouTube. The dataset was carefully curated with 25 indoor and 25 outdoor sequences covering a large variety of scene types like sky, sea/water, road/ground, ice, theatre stage, sports arena, cage etc. Tracking targets include both rigid and deformable/articulated objects like vehicle (motorcycle, car, bicycle), person, face, animal (fish, lion, puppies, birds, elephants, polar bear), aircraft (helicoptor, jet), boat and other generic objects (e.g sports ball). The application aspect was also kept into account while selecting the sequences, for example we include long sequences from theatre performances, music videos and movies, which are rich in content, and tracking in them may be useful in context of several recent applications like virtual camera simulation or video stabilization~\cite{kumar2017zooming,grundmann2011auto}. Similarly, long term tracking in sports videos can be quite helpful for automated analytics~\cite{lu2013learning}. The large variation in scene type and tracking targets can be observed in Figure \ref{fig:first_frames}. We further compare the TLP dataset with OTB in Figure \ref{fig:compare_OTB}, to highlight that the variation in bounding box size and aspect ratio with respect to the initial frame is significantly larger in TLP and the variations are also well balanced. The significant differences in duration of sequences in OTB and TLP are also apparent. 

The per sequence average length in TLP dataset is over 8 minutes. Each sequence is annotated with rectangular bounding boxes per frame, which were done using the VATIC~\cite{VATIC-2012}  toolbox. The annotation format is similar to OTB50 and OTB100 benchmarks to allow for easy integration with existing toolboxes. We have 33/50 sequences (amounting to 4\% frames in total) in TLP dataset where the target goes completely out of view and thus, we provide absent label for each frame in addition to the bounding box annotation. All the selected sequences are single shot (do not contain any cut) and have a resolution of $1280 \times 720$. Similar to VOT~\cite{kristan2015visual},  we choose the sequences without any cuts, to be empirically fair in evaluation, as most trackers do not explicitly model a re-detection policy. However, the recovery aspect of trackers still gets thoroughly evaluated on the TLP dataset, due to presence of full occlusions and out of view scenarios in several sequences. 
\paragraph{\bf TinyTLP and TLPattr:} We further derive two short sequence datasets from TLP dataset. The TinyTLP dataset consists of first 600 frames (20 sec) in each sequence of the TLP dataset to compare and highlight the challenges incurred due to long-term tracking aspect. The length of 20 sec was chosen to align the average per sequence length with OTB100 benchmark. The TLPattr dataset consists of total 90 short sequences focusing on different attributes. Six different attributes were considered in our work i.e (a) fast motion of target object or camera, (b) illumination variation around target object between consecutive frames, (c) large scale variation of the target object, (d) partial occlusions of the target object by other objects or background, (e) out of view or full occlusions, where object leaves the camera view or it is not visible at all and (f) background clutter. The TLPattr dataset includes 15 short sequences corresponding to each of the attribute. 

Each sequence in TLPattr is carefully selected in such a way that the only dominant challenge present in it is a particular attribute, it is assigned to. For example, for fast motion, we first select all instances in entire TLP dataset where the motion of the center of the ground truth bounding box between consecutive frames is more than 20 pixels. We temporally locate every such fast motion event and curate a short sequence around it by selecting 100 frames before and after the fast motion event. We then sort the short sequences based on the amount of motion (with the instance with most movement between two frames as the top sequence) and manually shortlist 15 sequences (starting from the top), where fast motion is the only dominant challenge present and simultaneously avoiding selection of multiple short sequences from the same long video. For attributes like illumination variation and background clutter the selection was fully manual. The rationale behind curating the TLPattr dataset was the following: (a) Giving a single attribute to entire sequence (as in previous works like OTB50) is ill posed on long sequences as in TLP. Any attribute based analysis with such an annotation would not capture the correct correlation between the challenge and the performance of the tracking algorithm. (b) Using per frame annotation of attributes is also difficult for analysis in long videos, as the tracker may often fail before reaching the particular frame where attribute is present and (c) The long sequences and variety present in TLP dataset allows us to single out a particular attribute and choose subsequences where that is the only dominant challenge. This paves the way for accurate attribute wise analysis. 

\section{Evaluation}
\label{section:eval}
\subsection{Evaluated Algorithms} 
We evaluated 17 recent trackers on the TLP and TinyTLP datasets. The trackers were selected based on three broad guidelines i.e.: (a) they are computationally efficient for large scale experiments; (b) their source codes are publicly available and (c) they are among the top performing trackers in existing benchmarks. Our list includes CF trackers with hand crafted features, namely SRDCF~\cite{danelljan2015learning}, MOSSE~\cite{bolme2010visual}, DCF~\cite{henriques2015high}, DSST~\cite{danelljan2014accurate},  KCF~\cite{henriques2015high}, SAMF~\cite{li2014scale}, Staple~\cite{Bertinetto_2016_CVPR}, BACF~\cite{kiani2017learning} and LCT~\cite{ma2015long}; CF trackers with with deep features: ECO~\cite{DanelljanCVPR2017} and CREST~\cite{song-iccv17-CREST} and deep trackers i.e. GOTURN~\cite{held2016learning}, MDNet~\cite{nam2016mdnet}, ADNet~\cite{yun2017adnet} and SiamFC\cite{bertinetto2016fully}. We also included TLD~\cite{kalal2012tracking} and MEEM~\cite{zhang2014meem} as two older trackers based on PN learning and SVM ensemble, as they specifically target the drift problem for long-term applications. We use default parameters on the publicly available version of the code when evaluating all the tracking algorithms.
\subsection{Evaluation Methodology} 
We use precision plot, success plot and longest subsequence measure for evaluating the algorithms. The precision plot~\cite{babenko2011robust,wu2013online} shows the percentage of frames whose estimated location is within the given threshold distance of the ground truth. A representative score per tracker is computed, by fixing a threshold over the distance (we use the threshold as 20 pixels). The success metric~\cite{wu2013online} computes the intersection over union (IoU) of predicted and ground truth bounding boxes and counts the number of successful frames whose IoU is larger than a given threshold. In out of view scenarios, if the tracking algorithm explicitly predicts the absence, we give it an overlap of 1 otherwise 0. The success plot shows the ratio of successful frames as the IoU threshold is varied from 0 to 1. A representative score for ranking the trackers is computed as the area under curve (AUC) of its success plot. We also employ the conventional success rate measure, counting frames above the threshold of 0.50 (IoU $>$ 0.50). 

\paragraph{\bf LSM metric:}We further propose a new metric called Longest Subsequence Measure (LSM) to quantify the long term tracking behaviour. The LSM metric computes the ratio of the length of the longest successfully tracked continuous subsequence to the total length of the sequence. A subsequence is marked as successfully tracked, if $x$\% of frames within it have IoU $>$ 0.5, where $x$ is a parameter. LSM plot shows the variation in the normalized length of longest tracked subsequence per sequence, as $x$ is varied. A representative score per tracker can be computed by fixing the parameter $x$ (we use the threshold as 0.95). 

The LSM metric captures the ability of a tracker to track continuously in a sequence within a certain bound on failure tolerance (parameter $x$) and bridges the gap over existing metrics which fail to address the issue of frequent momentary failures. For example, it often happens in long sequences that tracker loses the target at some location and freezes there. If coincidentally the target passes the same location (after a while), the tracker starts tracking it again. LSM penalizes such scenarios by considering only the longest continuous tracked subsequences.

\subsection{Per Tracker Evaluation}
\begin{figure}[t]
\centering
\tiny
 \scalebox{0.8}{\begin{tabular}{lccccccccccccccccc}
    \hline
     & MDNet & SiamFC & CREST & ADNet & GOTURN & ECO & MEEM & BACF & TLD & SRDCF & STAPLE & SAMF & DSST & LCT & DCF & KCF & MOSSE\\ \hline
        TinyTLP  & 83.4 & 70.1 & 65.8 & 68.7 & 51.8 & 57.6 & 49.2 & 60.0 & 36.4& 54.1 & 60.0 & 58.9 & 56.5 & 42.7 & 41.6 & 41.3 & 37.2\\
        TLP  & 42.1 & 27.6 & 24.9 & 22.1 & 22.0 & 21.8 & 19.5 & 15.9 & 13.8 & 13.2 & 13.1 & 11.3 & 8.8 & 8.7 & 7.9 & 6.9 & 3.7\\
    \hline
    \\
    \multicolumn{18}{c}{\includegraphics[width=13cm]{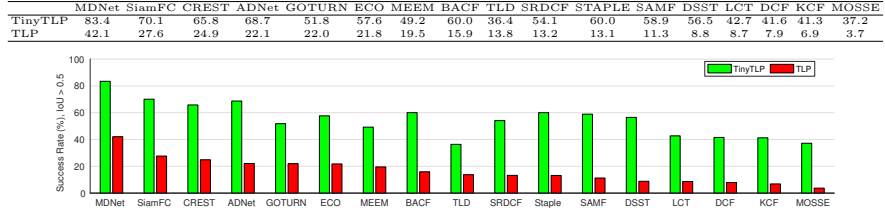}}  \end{tabular}} 

\label{fig:individual_results}
\caption{Success rate of individual trackers on TinyTLP and TLP datasets. The algorithms are sorted based on their performance on TLP.}
\end{figure}

Table~\ref{fig:individual_results} presents the success rate of each individual tracker on TinyTLP and TLP datasets. The MDNet tracker is the best performing tracker on both the datasets. TLD is the worst performing tracker on TinyTLP and MOSSE performs worst on TLP dataset. The performance significantly drops for each tracker on TLP dataset, when compared to TinyTLP dataset, which clearly brings out the challenges incurred in long-term tracking. The relative performance drop is minimum in MDNet where the success rate reduces from 83.4\% to 42.1\% (roughly by a factor of 2) and is most in MOSSE tracker, which reduces from 37.2\% in TinyTLP to 3.7\% in TLP (reduction by more than a factor of 10). 

In general, the relative performance decrease is more in CF trackers with hand crafted features as compared to CF+deep trackers. For instance, trackers like BACF, SAMF, Staple give competitive or even better performance than CREST and ECO over TinyTLP dataset, however, their performance steeply decreases on TLP dataset. Although all the CF based trackers (hand crafted or CNN based) are quite susceptible to challenges such as long term occlusions or fast appearance changes, our experiments suggest that using learnt deep features reduces accumulation of error over time and reduces drift. Such accumulation of error is difficult to quantify in short sequences and the performance comparison may not reflect the true ability of the tracker. For example, BACF outperforms ECO on TinyTLP by about 2\%, however it is 6\% worse than ECO on TLP. Similarly, the performance difference of SAMF and ECO is imperceptible on TinyTLP, which differs by almost a factor of 2 on TLP. 

The deep trackers outperform other trackers on TLP dataset, with MDNet and SiamFC being the top performing ones. ADNet is third best tracker on TinyTLP, however, its performance significantly degrades on TLP dataset. It is interesting to observe that both MDNet and ADNet refine last fully connected layer during online tracking phase, however, MDNet appears to be more consistent and considerably outperforms ADNet on TLP. The offline trained and freezed SiamFC and GOTURN perform relatively well (both appearing in top five trackers on TLP), however SiamFC outperforms GOTURN, possibly because it is trained on larger amount of video data. Another important observation is that the performance of MEEM surpasses all state of the art CF trackers with hand crafted features on TLP dataset. The ability to recover from failures also allows TLD tracker (giving lowest accuracy on TinyTLP) to outperform several recent CF trackers on TLP.

\subsection{Overall Performance}
\begin{figure}[t]
        \centering
        \begin{tabular}[b]{c c c}
        \includegraphics[width=4cm,height=3.2cm]{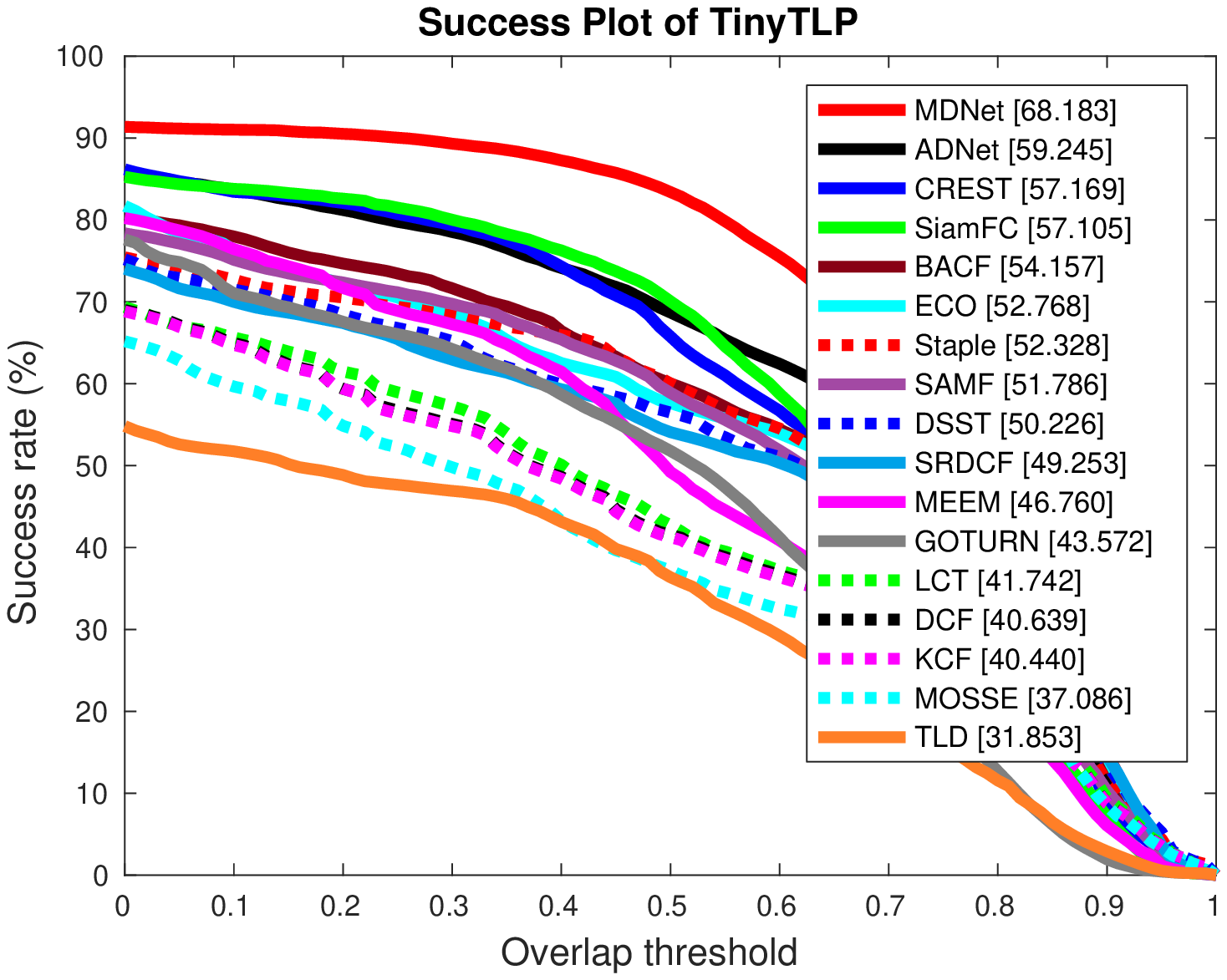}& 
        \includegraphics[width=4cm,height=3.2cm]{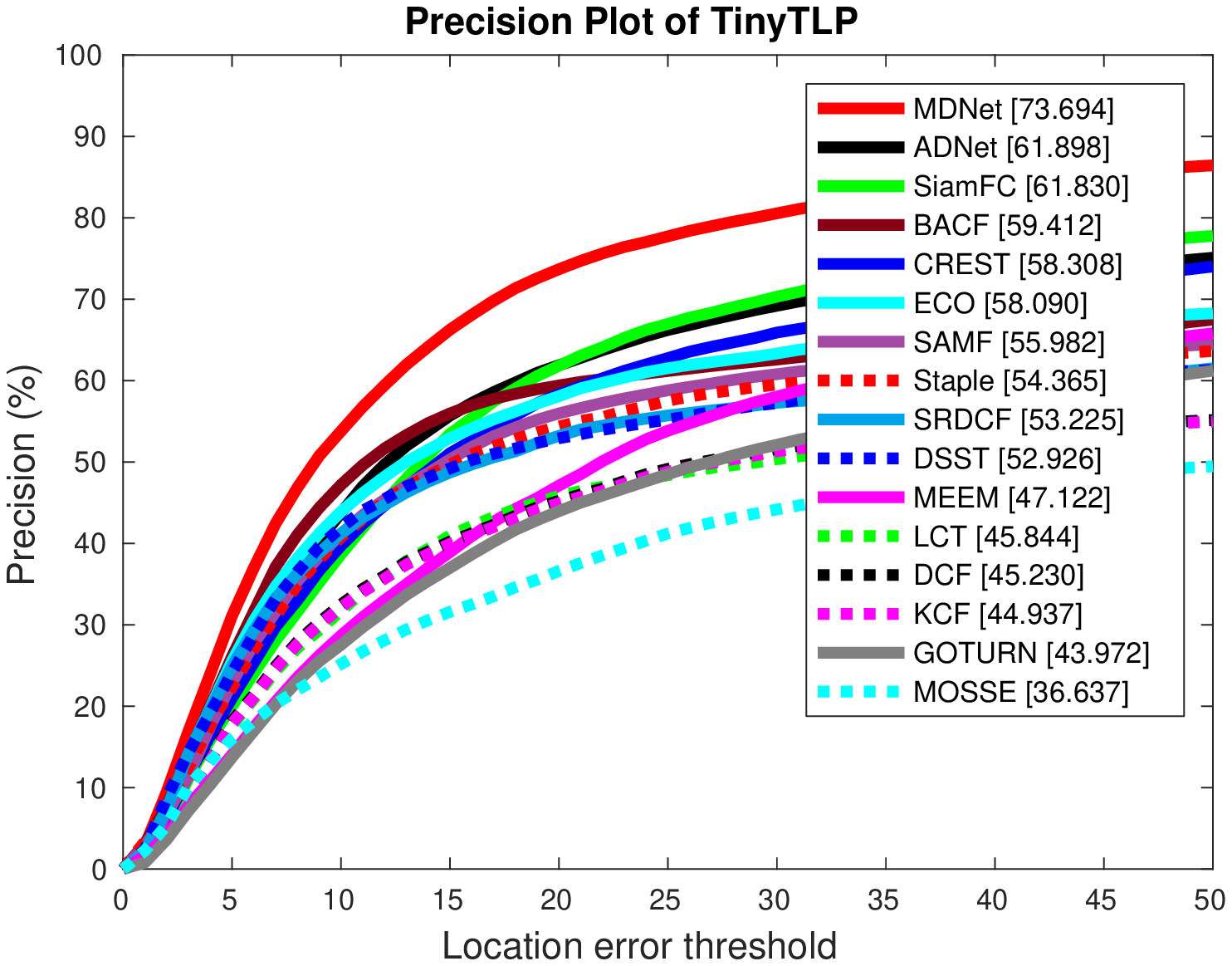}&
        \includegraphics[width=4cm,height=3.2cm]{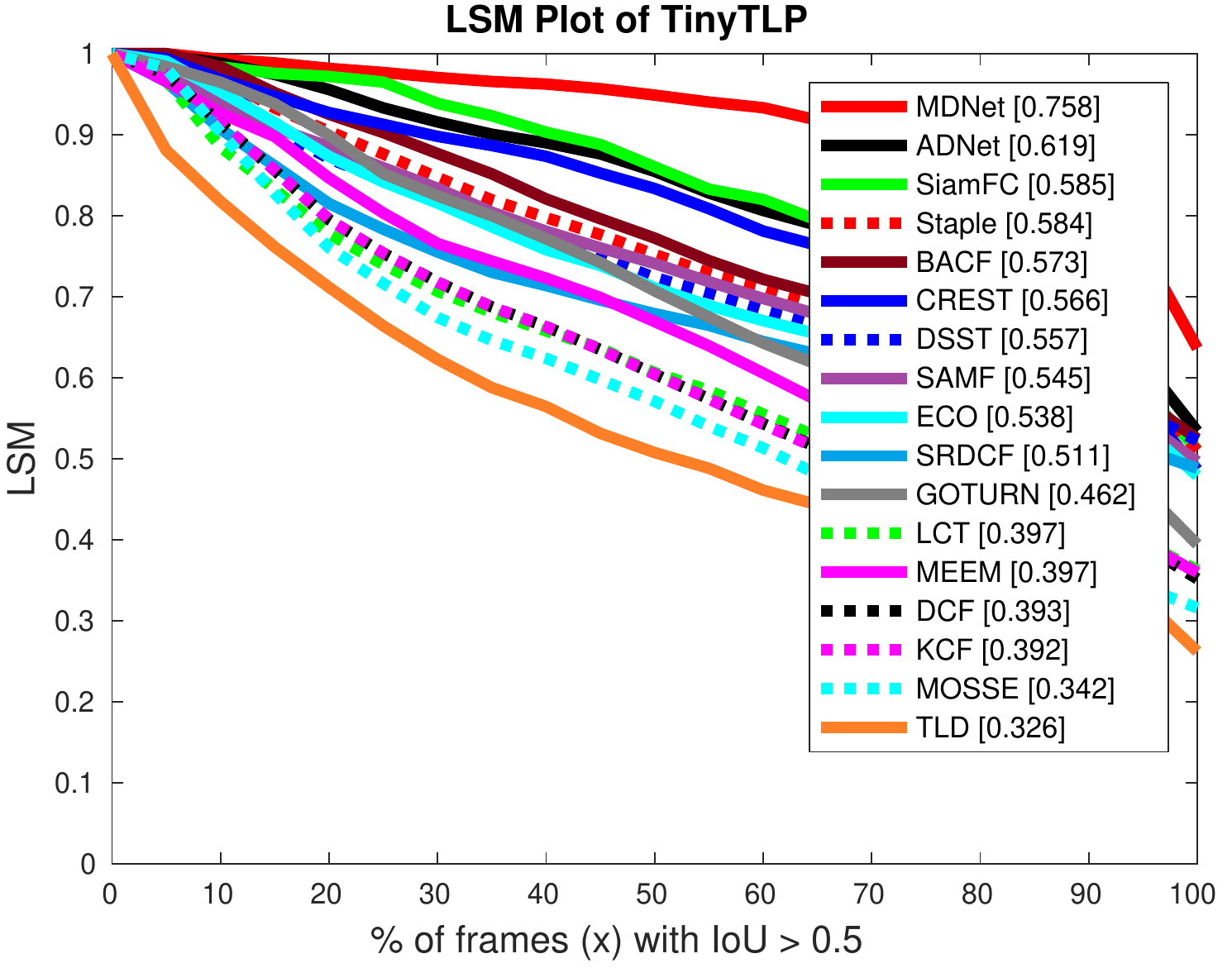} \\
        \includegraphics[width=4cm,height=3.2cm]{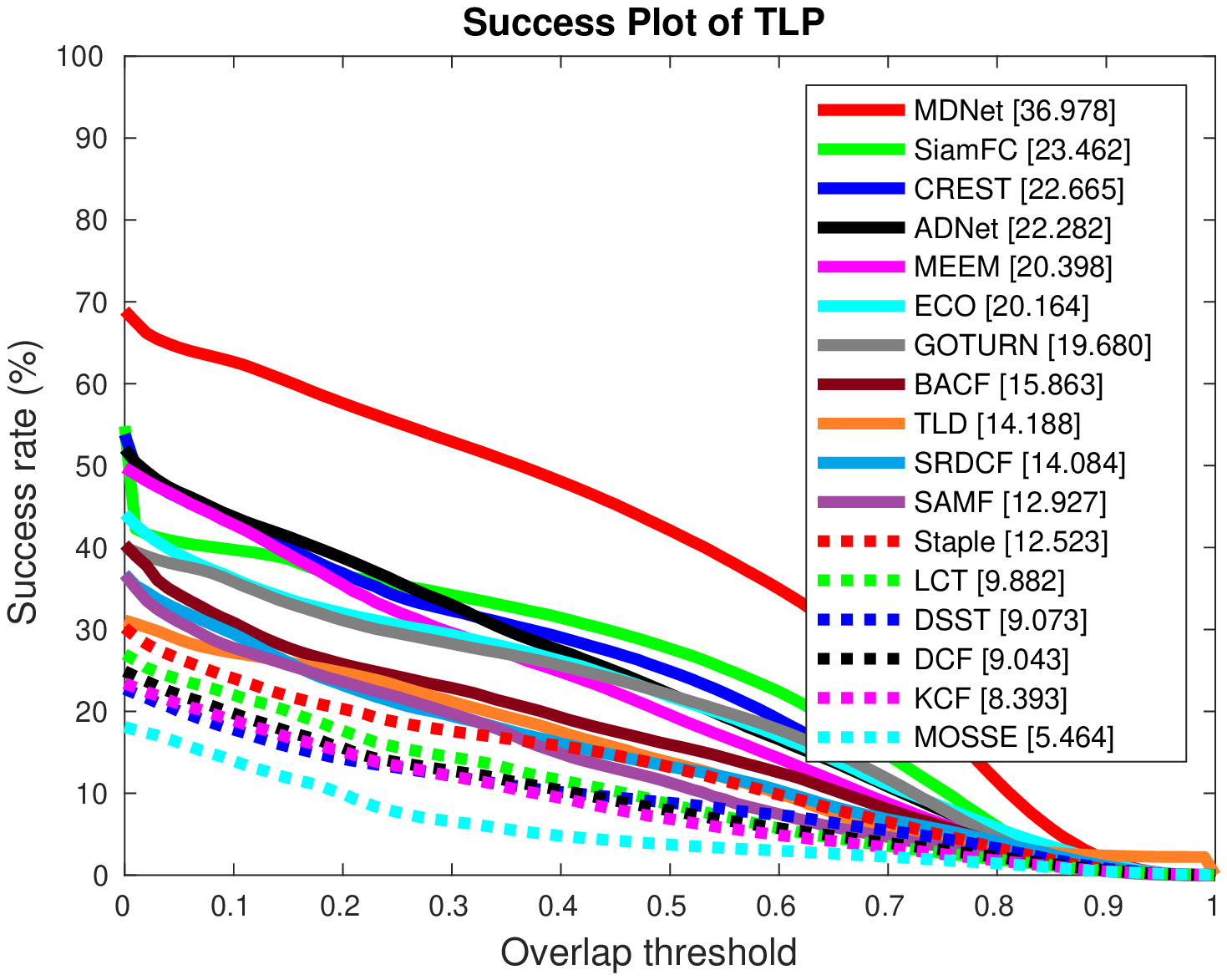}& 
        \includegraphics[width=4cm,height=3.2cm]{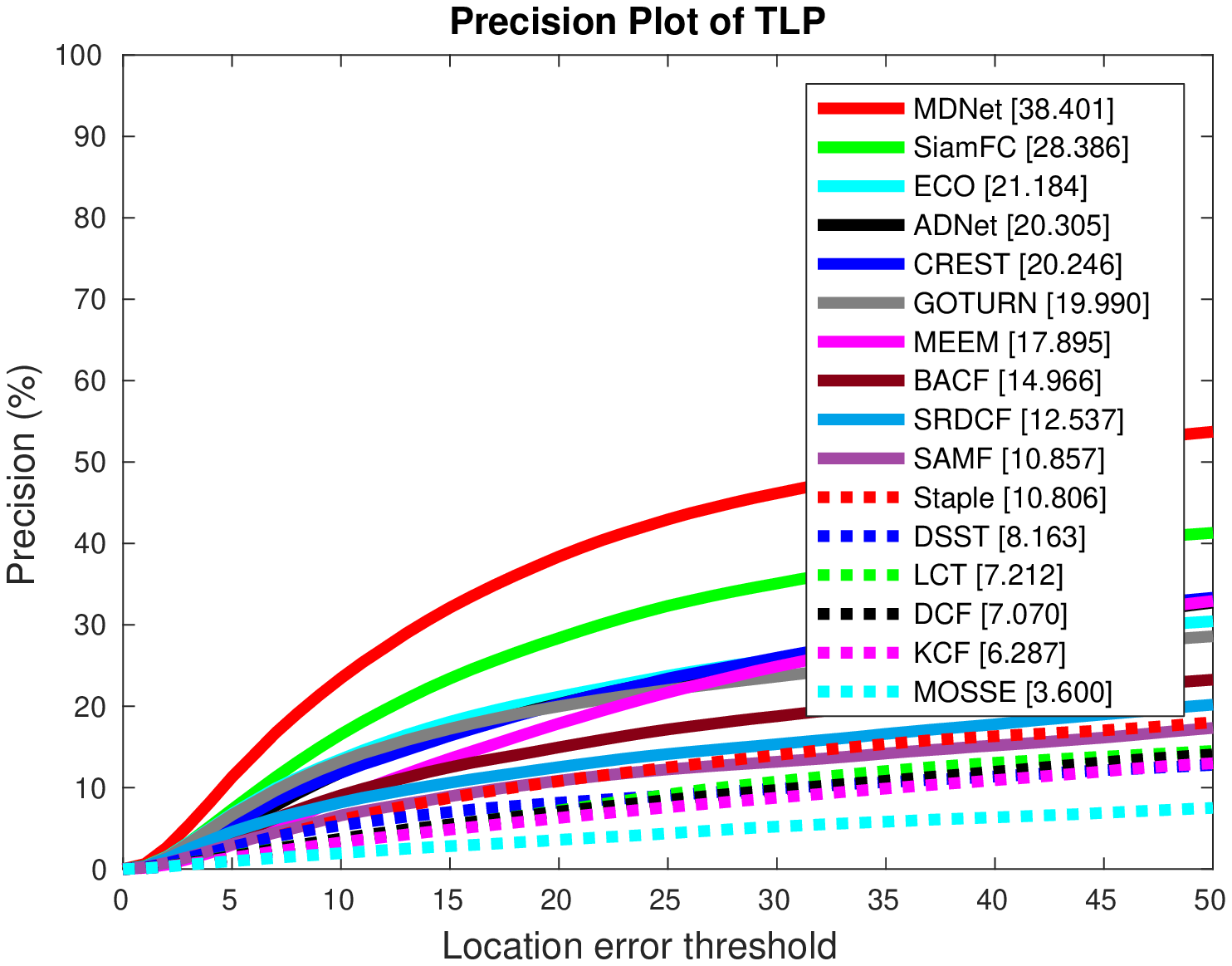}&
        \includegraphics[width=4cm,height=3.2cm]{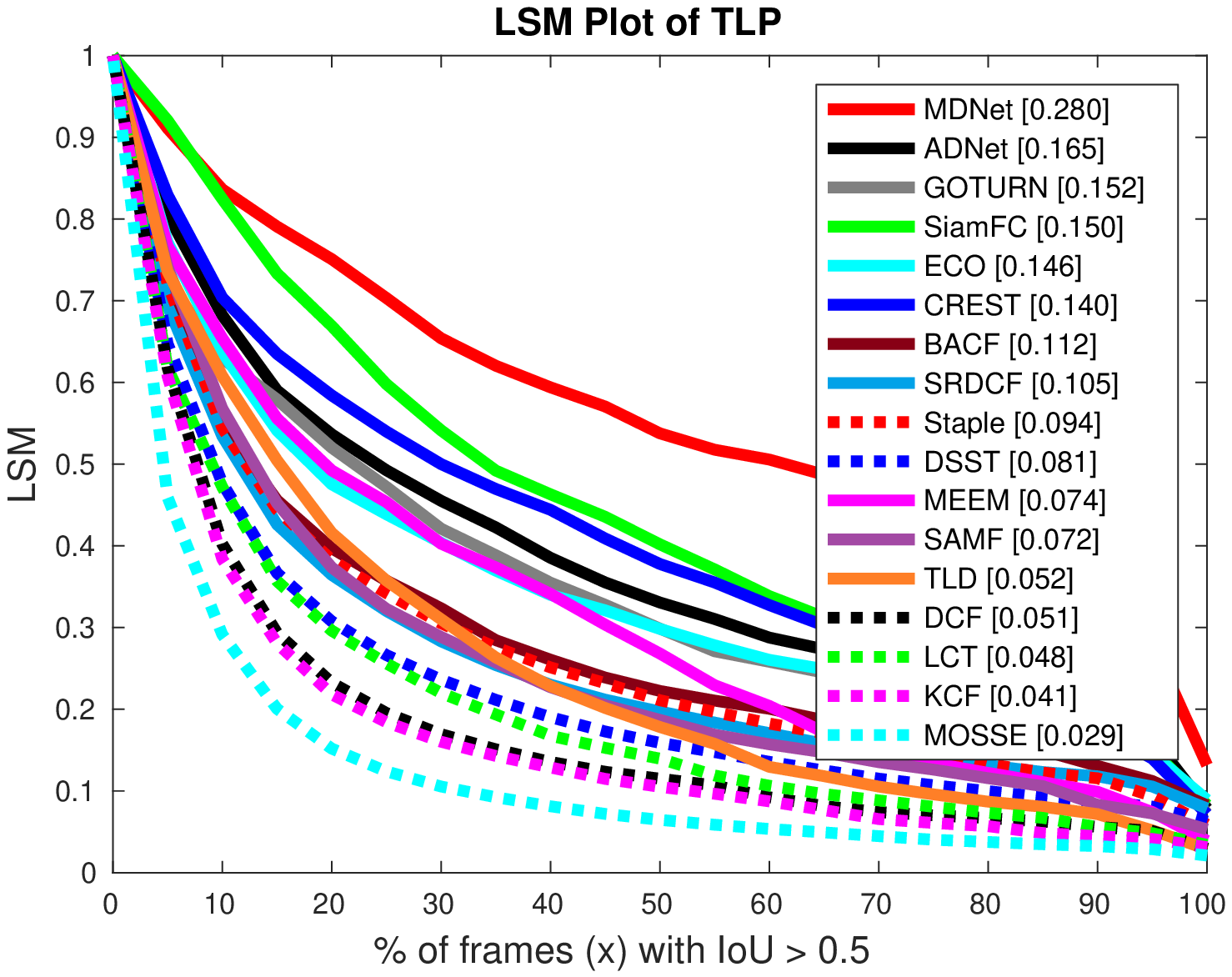} \\        
        \end{tabular}
\label{fig:overall_performance}
\caption{Overall Performance of evaluated trackers on TinyTLP and TLP with success plot, precision plot and LSM plot respectively (each column). For each plot, ranked trackers are shown with corresponding representative measure i.e. AUC in success plots; 20 pixel threshold in precision plots and 0.95 as length ratio in LSM plots.}
\end{figure}

The overall comparison of all trackers on TinyTLP and TLP using Success plot, Precision plot and LSM plot are demonstrated in Figure~\ref{fig:overall_performance}. In success plots, MDNet clearly outperforms all the other trackers on both TinyTLP and TLP datasets with AUC measure of 68.1\% and 36.9\% respectively. It is also interesting to observe that the performance gap significantly widens up on TLP and MDNet clearly stands out from all other algorithms. This suggests that the idea of separating domain specific information during training and online fine tuning of background and foreground specific information, turns out to be an extremely important one for long term tracking. Furthermore, analyzing MDNet and ADNet both of which employ the strategy of online updates on last FC layers during tracking, it appears that learning to detect instead of learning to track gives a more robust performance in long sequences. The performance drop of SiamFC and GOTURN on TLP also suggests a similar hypothesis.

The steeper success plots in TLP as compared to TinyTLP dataset, suggest that accurate tracking gets more and more difficult in longer sequences, possibly due to accumulation of error. The lower beginning point on TLP (around 40-50\% for most trackers compared to 80-90\% on TinyTLP), indicates that most trackers entirely drift away before reaching halfway through the sequence. The rankings in success plot on TLP are also quite contrasting to previous benchmarks. For instance, ECO is the best performing tracker on OTB100 closely followed by MDNet (with almost imperceptible difference), and its performance significantly slides on TLP. Interestingly, MEEM breaks into top five trackers in AUC measure of success plot on TLP (ahead of ECO). In general there is striking drop of performance between TinyTLP and TLP for most CF based trackers (more so for hand crafted ones). CREST is most consistent among them and ranks in top 5 trackers for both TinyTLP and TLP.

The precision plots also demonstrate similar trends as success plots, however they bring couple of additional subtle and interesting perspectives. The first observation is that SiamFC's performance moves closer to performance of MDNet on TLP dataset. Since SiamFC is fully trained offline and does not make any online updates, it is not accurate in scaling the bounding box to the target in long term, which brings down its performance in IoU measure. However, it still hangs on to the target due to the large scale training to handle challenges in predicting the consecutive bounding boxes, hence the numbers improve in the precision plot (again precision plot on TinyTLP does not capture this observation). The ADNet tracker is ranked two on TinyTLP using precision measure, however, it drops to 4th position on TLP. The GOTURN tracker also brings minor relative improvement in precision measure and moves ahead of MEEM on TLP. 

The LSM plots show the ratio of longest successfully tracked continuous subsequence to the total length of the sequence. The ratios are finally averaged over all the sequences for each tracker. A sequence is successfully tracked if $x$\% of frames in it have IoU $>$ 0.5. We vary the value $x$ to draw the plots and the representative number is computed by keeping $x=95$\%. This measure explicitly quantifies the ability to continuously track without failure. MDNet performs the best on this measure as well. The relative performance of CREST drops in LSM measure, as it partially drifts away quite often, however is able to recover from it as well. So its overall success rate is higher, however, the average length of longest continuous set of frames it can track in a video is relatively low. In general, the ratio of largest continuously tracked subsequence to sequence length (with success rate $>$ 0.95) averaged over all sequences is about 1/4th for MDNet and lower than 1/6th for other trackers. This indicates the challenge in continuous accurate tracking without failures.

\subsection{Attribute wise Performance Evaluation}
\begin{figure}[t]
        \centering
        \begin{tabular}[b]{c c c}
        {Illumination variation} & {Scale variation} & {Fast motion}\\
        \includegraphics[width=0.32\linewidth]{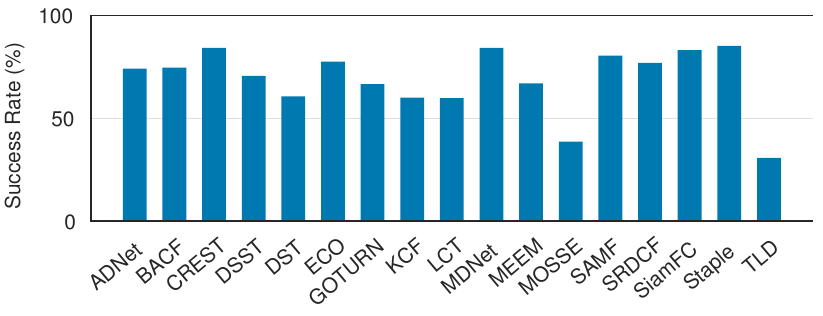}& 
        \includegraphics[width=0.32\linewidth]{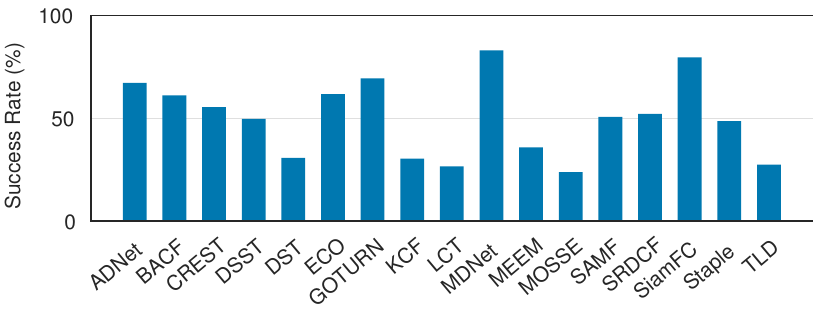}& 
        \includegraphics[width=0.32\linewidth]{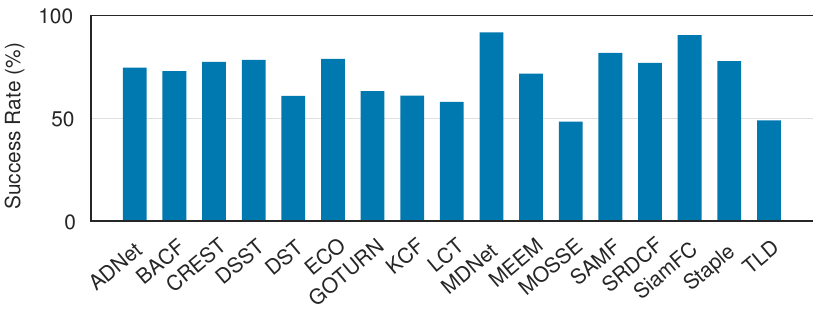} \\
        {Background Clutter} & {Partial occlusions} & {Out of view}\\
        \includegraphics[width=0.32\linewidth]{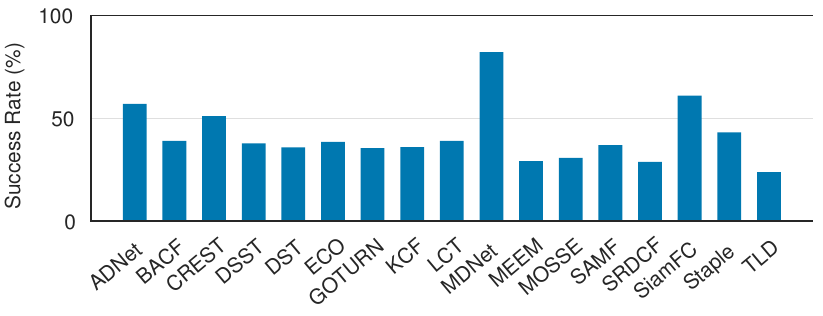}& 
        \includegraphics[width=0.32\linewidth]{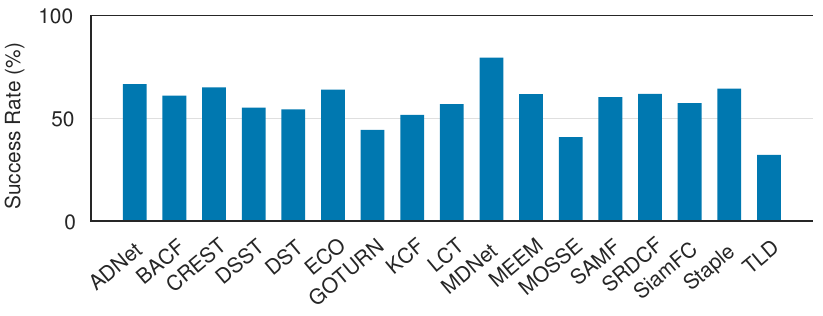}& 
        \includegraphics[width=0.32\linewidth]{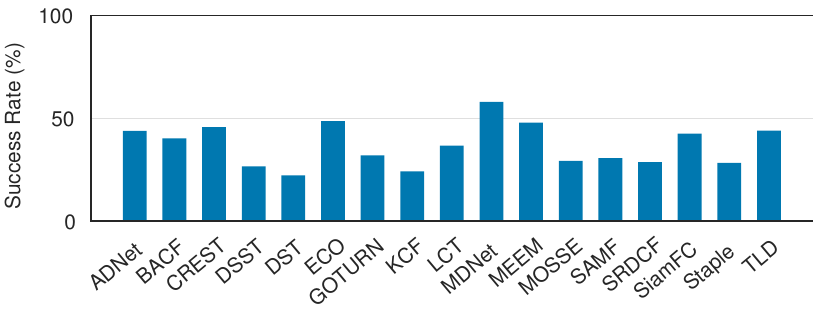} \\  
        \end{tabular}
        \caption{Attribute wise performance evaluation on TLPattr dataset. Results are reported as success rate (\%) with IoU $>$ 0.5. }
        \label{fig:attribute_wise_performance}
\end{figure}
The average attribute wise success rates of all the trackers on TLPattr dataset are shown in Figure~\ref{fig:attribute_wise_performance}. Each attribute in TLPattr dataset includes 15 short sequences corresponding to it (dominantly representing the particular challenge). Out of view appears to be the most difficult challenge hindering the performance of the trackers followed by background clutter, scale variation and partial occlusions. Most of the trackers seem to perform relatively better on sequences with illumination variation and fast motion. On individual tracker wise comparison, MDNet gives best performance across all the attributes, clearly indicating the tracker's reliable performance across different challenges. 

Another important perspective to draw from this experiment is that the analysis on short sequences (even if extremely challenging) is still not a clear indicator of their performance on long videos. For example, Staple and CREST are competitive in performance across all the attributes, however their performance on full TLP dataset differs by almost a factor of two in success rate measure (CREST giving a value 24.9 and Staple is only 13.1). Similarly comparison can be drawn between DSST and GOTURN, which are competitive in per attribute evaluation (with DSST performing better than GOTURN on fast motion, partial occlusions, background clutter and illumination variation). However, in long terms setting, their performance varies by a large margin (GOTURN giving success rate of 22.0, while DSST is much inferior with a value of 8.8).

\subsection{Evaluation on repeated TinyTLP sequences}
\begin{figure}[t]
        \centering
        \begin{tabular}[b]{c c c}
        \includegraphics[width=0.3\linewidth]{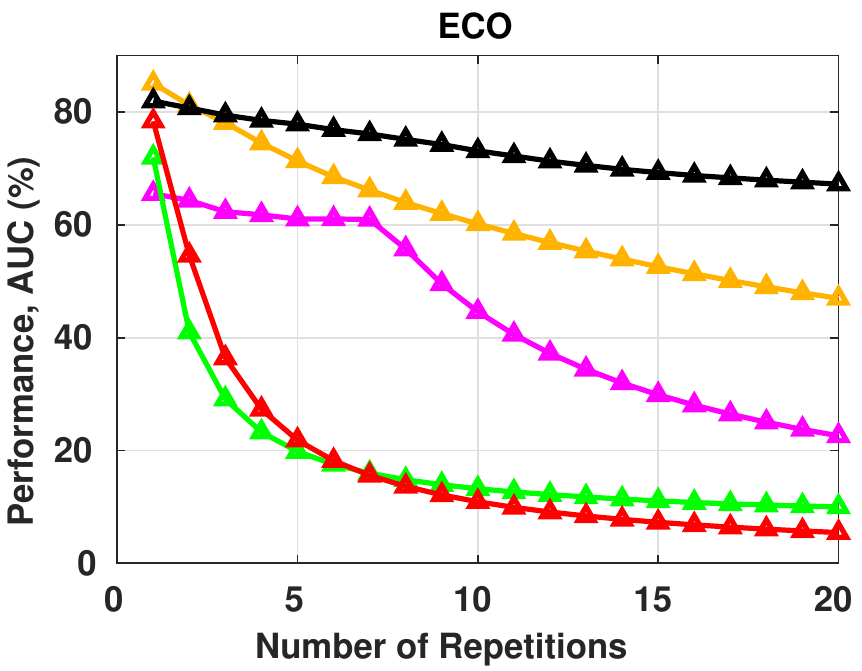}& 
        \includegraphics[width=0.3\linewidth]{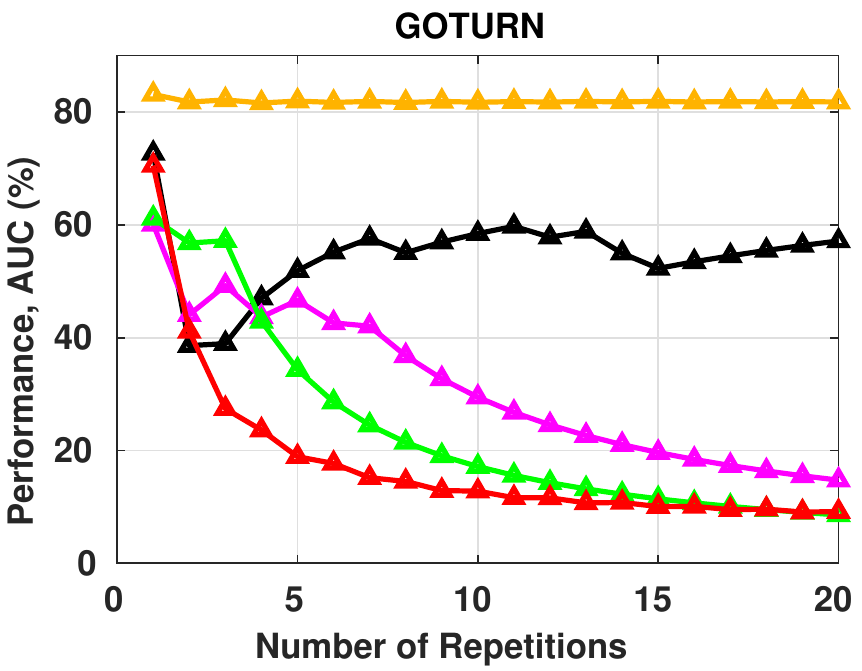}& 
        \includegraphics[width=0.3\linewidth]{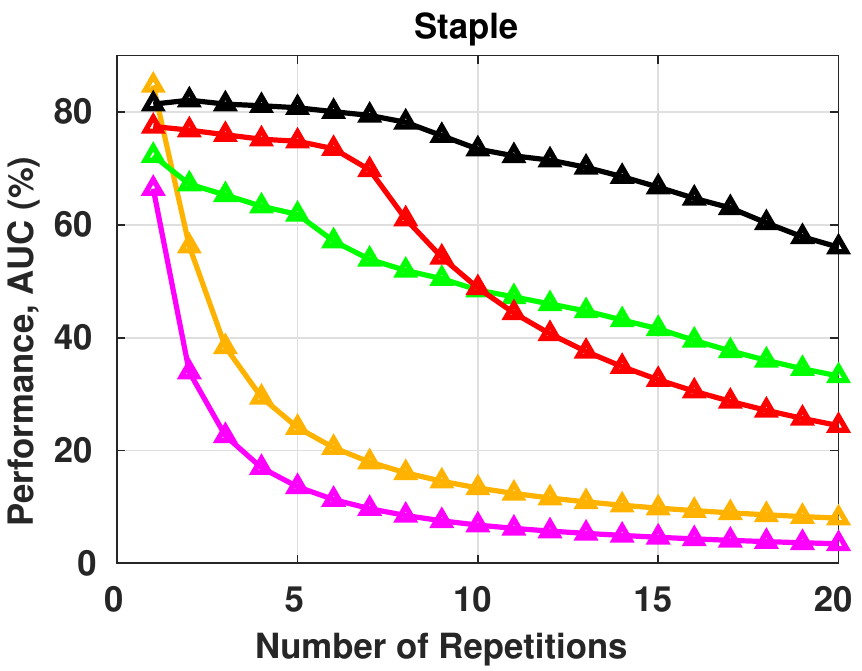} \\
        \end{tabular}
        \caption{Results of three different trackers on 20 times elongated TinyTLP sequences (by reversing and concatenating the sequence in iterative way). Each color represents a different sequence and each triangle represents a repetition.}
        \label{fig:repeat_experiment}
\end{figure}

The essence of our paper is the need to think ``long term" in object tracking, which is crucial for most practical applications of tracking. However, it remains unclear if there exists a ``long term challenge in itself" and one can always argue that the performance drop in long videos is just because of ``more challenges" or ``frequent challenges". To investigate this further, we conduct a small experiment where we take a short sequence and repeat it 20 times to make a longer video out of it, by iteratively reversing and attaching it at the end to maintain the continuity. This increases the length of the sequence without introducing any new difficulty or challenges. In Figure~\ref{fig:repeat_experiment}, we present such an experiment with three different trackers ECO (deep+CF, best performing tracker on OTB), GOTURN (pure deep) and Staple (pure CF) on 5 TinyTLP sequences for each tracker, where the tracker performs extremely well in the first iteration. We can observe that the tracking performance degrades for all three algorithms (either gradually or steeply) as the sequences get longer, which occurs possibly due to error accumulated over time. This again highlights the fact the tracking performance not just depends on the challenges present in the sequence but also gets affected by the length of the video. Hence, a dataset like TLP, interleaving the challenges and the long term aspect, is necessary for comprehensive evaluation of tracking algorithms.

\subsection{Run time comparisons}
\begin{figure}[t]
        \centering
        \includegraphics[width=\textwidth]{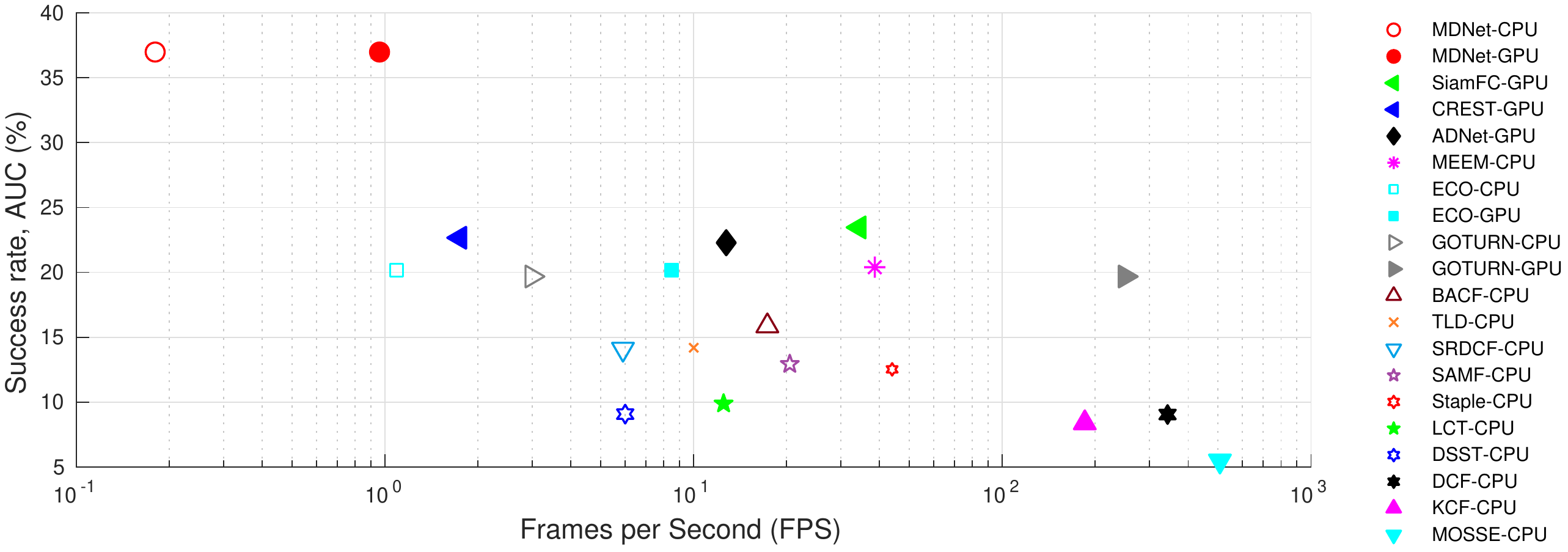} 
         \caption{Runtime comparison of different tracking algorithms.}
        \label{fig:runtime_performance}
\end{figure}

The run time speeds of all the evaluated algorithms are presented in Figure~\ref{fig:runtime_performance}. For fair evaluation, we tested all the CPU algorithms on a 2.4GHz Intel Xeon CPU with 32GB RAM and we use a NVIDIA GeForce GTX 1080 Ti GPU for testing GPU algorithms. The CF based trackers clearly are most computationally efficient and even CPU algorithms run several folds faster than real time. The deep CF and deep trackers are computationally more expensive. MDNet gives lowest tracking speeds and runs at 1 FPS even on GPU. Among deep trackers GOTURN is the fastest tracker, however SiamFC and ADNet bring a good trade off in terms of overall success rate and run time speeds on GPU. 
\section{Conclusion}
This work aims to emphasize the fact that tracking on large number of tiny sequences, does not clearly bring out the competence or potential of a tracking algorithm. Moreover, even if a tracking algorithm works well on extremely challenging small sequences and fails on moderately difficult long sequences, it will be of limited practical importance. To this end, we propose the TLP dataset, focusing on the long term tracking application, with notably larger average duration per sequence, a factor which is of extreme importance and has been neglected in the existing benchmarks. We evaluate 17 state of the art algorithms on the TLP dataset, and the results clearly demonstrate that almost all state of the art tracking algorithms do not generalize well on long sequence tracking, MDNet being the only algorithm achieving more than 25\% on the AUC measure of success plots. However, MDNet is also the slowest among the evaluated 17 trackers in terms of run time speeds.

Interestingly, if we only select the first 20 seconds of each sequence for evaluation (calling it TinyTLP dataset), the performance of all the trackers increases by multiple folds across different metrics. Another important observation is that the evaluations on small datasets fail to efficiently discriminate the performances of different tracking algorithms, and closely competing algorithms on TinyTLP result in quite different performance on TLP. The dominant performance of MDNet suggests that the ideas of online updating the domain specific knowledge and learning a classifier cum detector instead of a tracker (which regresses the shift), are possibly some cues to improve the performance in long term setting. Our evaluation on repeated TinyTLP sequences shows that temporal depth indeed plays an important role in the performance of evaluated trackers and appropriately brings out their strengths and weaknesses. To the best of our knowledge, TLP benchmark is the first large-scale evaluation of the state of the art trackers, focusing on long duration aspect and makes a strong case for much needed research efforts in this direction, in order to track long and prosper. 
%
%
\bibliographystyle{splncs04}
\bibliography{egbib}
\end{document}